\documentclass[iicol, pdflatex,sn-mathphys-num]{sn-jnl}

\usepackage{graphicx}%
\usepackage{multirow}%
\usepackage{amsmath,amssymb,amsfonts}%
\usepackage{amsthm}%
\usepackage{mathrsfs}%
\usepackage[title]{appendix}%
\usepackage{xcolor}%
\usepackage{textcomp}%
\usepackage{manyfoot}%
\usepackage{booktabs}%
\usepackage{algorithm}%
\usepackage{algorithmicx}%
\usepackage{algpseudocode}%
\usepackage{listings}

\usepackage{comment}
\usepackage{siunitx}
\usepackage{xurl}%
\usepackage{url}
\usepackage{makecell}
\usepackage{threeparttable}
\usepackage{tabularx}

\usepackage{tikz}
\usetikzlibrary{trees, positioning}
\usepackage{caption}

\newcounter{promptcounter}
\newcommand{\promptcaption}[1]{%
    \refstepcounter{promptcounter}%
    \raggedright \textbf{Prompt \thepromptcounter:} #1
}

\theoremstyle{thmstyleone}%
\theoremstyle{thmstyletwo}%

\theoremstyle{thmstylethree}%

\begin{document}

\title[Article Title]{Toward Purpose-oriented Topic Model Evaluation enabled by Large Language Models}


\author*[1]{\fnm{Zhiyin} \sur{Tan}}\email{zhiyin.tan@l3s.de}

\author[2]{\fnm{Jennifer} \sur{D’Souza}}\email{jennifer.dsouza@tib.eu}

\affil*[1]{\orgdiv{L3S Research Center}, \orgname{Leibniz University Hannover}, \orgaddress{\street{Appelstraße 9a}, \city{Hannover}, \postcode{30167}, \state{Lower Saxony}, \country{Germany}}}

\affil[2]{\orgdiv{Data Science \& Digital Libraries}, \orgname{TIB Leibniz Information Centre for Science and Technology}, \orgaddress{\street{Welfengarten 1 B}, \city{Hannover}, \postcode{30167}, \state{Lower Saxony}, \country{Germany}}}


\abstract{This study presents a framework for automated evaluation of dynamically evolving topic models using Large Language Models (LLMs). Topic modeling is essential for organizing and retrieving scholarly content in digital library systems, helping users navigate complex and evolving knowledge domains. However, widely used automated metrics, such as coherence and diversity, often capture only narrow statistical patterns and fail to explain semantic failures in practice. We introduce a purpose-oriented evaluation framework that employs nine LLM-based metrics spanning four key dimensions of topic quality: lexical validity, intra-topic semantic soundness, inter-topic structural soundness, and document-topic alignment soundness. The framework is validated through adversarial and sampling-based protocols, and is applied across datasets spanning news articles, scholarly publications, and social media posts, as well as multiple topic modeling methods and open-source LLMs. Our analysis shows that LLM-based metrics provide interpretable, robust, and task-relevant assessments, uncovering critical weaknesses in topic models such as redundancy and semantic drift, which are often missed by traditional metrics. These results support the development of scalable, fine-grained evaluation tools for maintaining topic relevance in dynamic datasets.}


\keywords{Topic Modeling, Evaluation Framework, Large Language Models (LLMs)}



\maketitle

\section{Introduction}\label{sec1}

Topic modeling is a foundational technique for uncovering latent thematic structures in large-scale text corpora. By organizing unstructured content into interpretable topic clusters, topic models support a wide array of tasks, such as document organization~\cite{blei2003LDA}, content recommendation~\cite{wang2011CTR}, and longitudinal trend analysis~\cite{dwivedi-2023-evolutionAI}. These capabilities are especially valuable in scholarly and technical information systems, where users navigate vast and diverse literature under evolving research paradigms. In such settings, topic models serve not only as exploratory tools but also as building blocks for dynamic taxonomies, helping to index, summarize, and connect scientific knowledge at scale. 

However, as research domains grow more complex and specialized, traditional human-centric and static evaluation pipelines often fail to scale or remain reliable. Existing automated metrics, such as topic coherence~\cite{roder2015CV}, topic diversity~\cite{dieng2020ETM}, capture only limited aspects of topic quality, offering little insight into why certain topics under-perform in downstream applications or fail to align with human expectations.

Recent advances in instruction-tuned large language models (LLMs) offer promising alternatives. With their contextual reasoning and linguistic fluency, LLMs have demonstrated potential in evaluating generated content~\cite{stammbach2023revisitingLLM}, including initial efforts in scoring topic coherence. Yet current LLM-based evaluations are ad hoc and fragmented, lacking a purpose-oriented structure or comprehensive validation across modeling paradigms, domains, and task scenarios.

In this paper, we introduces a unified and scalable evaluation framework that leverages LLMs to assess topic model outputs across multiple quality dimensions. Our framework is motivated by the diverse roles that topic models play in complex information environments, such as scholarly retrieval systems, where reliable, interpretable, and adaptive topic taxonomies are essential. 
We operationalize nine LLM-based metrics across different distinct dimensions: lexical validity of topic words, intra-topic semantic soundness, inter-topic structural soundness, and document-topic alignment soundness. Each dimension is grounded in concrete usage scenarios.  
We conduct a comprehensive empirical study involving three datasets (news groups, scientific abstracts, and social media), four representative topic modeling methods (probabilistic, neural, and clustering-based), and eight instruction-tuned LLMs, ranging from lightweight to large-scale architectures. 
Our contributions are threefold:  
\begin{enumerate} 
    \item We design a modular LLM-as-a-judge evaluation framework for topic modeling that aligns real-world application demands with interpretable quality criteria, each assessed by one or more dedicated metrics.  
    \item We propose adversarial and sampling-based testing protocols to examine metric reliability, robustness, and alignment with known failure cases.     
    \item We present a multi-dimensional analysis of topic model behavior, revealing how modeling choices affect trade-offs among key evaluation rubrics such as coherence, diversity, and alignment, and how different LLMs vary in evaluative and diagnostic power. 
\end{enumerate}  
Our findings provide new tools for scalable and purpose-aligned evaluation of topic models, advancing their use in settings where taxonomy quality and semantic precision are critical to user-facing applications.

{This paper is a substantially extended version of our work presented at the 21st Conference on Information and Research Science Connecting to Digital and Library Science (IRCDL 2025) \cite{Tan2025Bridging}. Compared to the conference version, this manuscript introduces four main contributions. First, we formalize our approach into a purpose-oriented evaluation framework that connects metrics to real-world applications. Second, we expand our methodology by introducing a new lexical validity dimension with corresponding metrics and tests. Third, we broaden our empirical study by adding a social media dataset (TWEETS\_NYR) and increasing the number of LLM evaluators from three to eight. Finally, we present new findings through detailed correlation analysis and a systematic investigation of LLM evaluator behavior, offering insights into metric relationships and resource-efficient model selection.}
The code and dataset are publicly available\footnote {\url{https://github.com/zhiyintan/topic-model-LLMjudgment}}.

\section{The Evolution of Topic Modeling and Evaluation Paradigms}\label{sec:evolution}

Topic modeling has evolved into a foundational technique for uncovering latent semantic structure in large textual corpora. While the design of topic models has advanced significantly over time, so too have the evaluation paradigms that accompany them. These developments are not independent: the metrics we use to assess topic models have co-evolved with modeling assumptions, technological capabilities, and shifting expectations of utility. This section traces the progression of topic modeling alongside its evaluation practices, highlighting how changes in model architecture have motivated new evaluation priorities.

A typical topic model takes as input a corpus of $N$ documents $\{d_1, d_2, \dots, d_N\}$ drawn from a fixed vocabulary of $V$ words. The model is configured to discover $K$ latent topics, each represented by a ranked list of $M$ topic words, denoted $\mathcal{T}k = \{w_1, w_2, \dots, w_M\}$ for topic $k$. Each document $d_i$ is assigned a topic distribution $\theta_i$, a length-$K$ vector where $\theta\{i,k\}$ reflects the degree of association between the document and topic $k$. These topic-word and document-topic associations jointly serve as the core outputs of a topic model.

Initial topic modeling techniques prioritized statistical fit and generative plausibility. Latent Semantic Indexing (LSI)~\cite{papadimitriou1998LSI} applied matrix factorization to word-document co-occurrence matrices, while Probabilistic Latent Semantic Indexing (pLSI)~\cite{hofmann1999pLSI} introduced probabilistic mixture modeling over topics. Latent Dirichlet Allocation (LDA)~\cite{blei2003LDA} established a fully generative framework with Dirichlet priors over $\theta_i$ and topic-word distributions $\phi_k$, enabling principled inference and generalization.

Evaluation in this phase focused on internal model performance: held-out log-likelihood, perplexity, and predictive accuracy~\cite{blei2003LDA, blei2007CTM, newman2009distributed}. These metrics quantified statistical generalizability, but did not reflect interpretability or downstream usability. As noted by \citet{chang2009HumanTeaLeaves}, better likelihood does not necessarily produce more coherent or useful topics.

To address the disconnect between statistical fit and semantic clarity, researchers introduced metrics grounded in human interpretability.
\citet{chang2009HumanTeaLeaves} proposed the word intrusion task, showing that high-likelihood models can yield incoherent topics. This motivated a wave of coherence-based metrics, including $C_\text{UMass}$~\cite{mimno2011optimizing}, $C_\text{UCI}$~\cite{Newman2010TMforDL}, $C_\text{NPMI}$~\cite{bouma2009NPMI}, and $C_\text{V}$~\cite{roder2015CV}. These metrics quantify intra-topic association via word co-occurrence statistics or normalized PMI.

Such measures have since been widely adopted to evaluate both probabilistic and neural models, e.g. ProdLDA~\cite{srivastava2017ProdLDA-AVITM}, SCHOLAR~\cite{card2018SCHOLAR}, ATM~\cite{wang2019ATM}, ETM~\cite{dieng2020ETM}, BAT~\cite{wang2020BAT}, CombinedTM~\cite{bianchi2021CombinedTM}, and ECRTM~\cite{wu2023ECRTM}, highlighting a growing emphasis on human-readable topics.

{Co-occurrence-based} coherence metrics struggle in sparse, multilingual, or domain-specific corpora where related words may not co-occur. Embedding-based approaches~\cite{schnabel2015EmbeddingEvalTM, Nikolenko2016EmbeddingEvalTM, ding2018coherenceNTM} address this by computing semantic similarity between topic words using pretrained embeddings.
Building on embedding representations, recent approaches such as \citet{sia2020tired-cluster} and BERTopic~\cite{grootendorst2022bertopic} explore clustering-based topic formation, further relaxing generative assumptions and enabling more flexible topic discovery pipelines.

In parallel, evaluation expanded to assess topic distinctiveness and structure. Metrics such as topic diversity~($D_\text{TD}$)~\cite{dieng2020ETM}, redundancy~($D_\text{TR}$)~\cite{Burkhardt2019TopicRedundancy}, uniqueness~($D_\text{TU}$)~\cite{nan2019TopicUniqueness}, and inverted rank-biased overlap~ ($D_\text{IRBO}$)\cite{bianchi2021CombinedTM} quantify lexical overlap across topics. Word embedding-aware extensions including word embedding-based centroid distance ($D_\text{WE-CD}$)\cite{bianchi2021EmbedingCentroidTD}, word embedding-based pairwise distance ($D_\text{WE-PD}$)~\cite{Terragni2021EmbedingTD}, and word embedding-based inverted rank-biased overlap~($D_\text{WE-IRBO}$)~\cite{Terragni2021EmbedingTD} further measure semantic separation.

These metrics reflect a shift in evaluation priorities: beyond coherence, users demand topic models that produce diverse, non-redundant, and structurally interpretable topic sets.

As topic models are increasingly used for indexing, retrieval, and summarization, new evaluation metrics have emerged to assess how well topics align with document content.
Early work, such as \citet{bhatia2017DocumentTopicEval} introduced human-annotated document–topic relevance and the topic intrusion task; \citet{bhatia2018topicIntrusionEval} later automated this process with a trained classifier. Clustering-based document evaluations~\cite{Korencic2018DocumenCoherence}, topic coverage measures~\cite{korenvcic2021CoverageEval}, contribute to this growing class of document-level evaluation metrics. These approaches emphasize that the utility of topic models depends not only on word-level semantics but also on document-level alignment.

More recently, LLM-based evaluations have emerged as a promising new paradigm. Studies such as \citet{stammbach2023revisitingLLM} and \citet{rahimi2024contextualizedCoherence} demonstrate that large language models can simulate human reasoning, providing nuanced judgments of topic coherence.
\citet{yang2024llmReadingTeaLeaves} proposes metrics to quantify the agreement between LLM-generated keywords from documents and topic model outputs. By leveraging LLMs’ world knowledge and contextual understanding, these methods address the limitations of traditional statistical metrics and reduce reliance on human evaluations.

{However, most current LLM-based evaluation methods focus on a single aspect, such as coherence or word intrusion, without fully addressing other important dimensions or the diverse requirements of real-world applications. Despite the proliferation of evaluation metrics, many remain model-internal or task-agnostic. As \citet{blei2012PLDA} noted, a structural disconnect persists between how we evaluate topic models and how we expect them to function. \citet{doogan-buntine-2021-TopicTwaddle} further argues that interpretability does not imply utility, highlighting the lack of purpose-aware design in evaluation.
A unified framework that leverages the strengths of LLMs to provide a comprehensive and interpretable evaluation, closely aligned with practical application scenarios, would therefore be highly beneficial.}

\definecolor{group}{HTML}{91A4C9}
\definecolor{groupbg}{HTML}{EDF2FC}
\definecolor{rate}{HTML}{A577B8}
\definecolor{ratebg}{HTML}{F1E4F6}
\definecolor{count}{HTML}{92B587}
\definecolor{countbg}{HTML}{F3FAF1}
  
\begin{figure*}[htbp]
\centering
\begin{tikzpicture}[
  grow=right, 
  sibling distance=20mm,
  level distance=60mm, 
  level 1/.style={level distance=47.5mm},
  level 2/.style={level distance=67mm},
  inner xsep=4.9pt, inner ysep=4pt,
  every node/.style={
    draw, rectangle, rounded corners, 
    align=left, anchor=center,
    minimum height=5mm, 
    font=\fontsize{8}{10}\selectfont,
    },
  solo/.style={text width=1.8cm, minimum height=15.5mm},
  double/.style={text width=1.8cm, minimum height=17.6mm},
  children1/.style={text width=5.8cm, minimum height=17mm},
  children2/.style={text width=6.3cm, minimum height=17mm},
  children-end/.style={text width=12.75cm, minimum height=10.5mm},
  edge from parent path={(\tikzparentnode.east) -- ++(3.4mm,0) |- (\tikzchildnode.west)},
  boxGroup/.style={draw=group, fill=groupbg},
  boxRate/.style={draw=rate, fill=ratebg},
  boxCount/.style={draw=count, fill=countbg}
]
\node[double, boxGroup]{Document\\-Topic Alignment Soundness}
  child [level distance=82.2mm, sibling distance=13.6mm] {node[children-end, boxCount] {\textbf{$A_{\text{ir-tw}}$ (count $\downarrow$)} [Sec~\ref{subsubsec:airtw}]\\Irrelevant Topic Words Detection: Finds document themes not covered by assigned topic words.}
  }
  child [level distance=82.2mm, sibling distance=13.6mm] {node[children-end, boxCount] {\textbf{$A_{\text{missing-theme}}$ (count $\downarrow$)} [Sec~\ref{subsubsec:amissingtheme}]\\Missing Theme Detection: Flags topic words irrelevant to an assigned document’s content.}
  };
\begin{scope}[yshift=2.03cm]
\node[solo, boxGroup]{Inter-topic Structural Soundness}
  child [level distance=82.2mm] {node[children-end, boxRate] {\textbf{$D_{\text{rate}}$ (rate 1-3 $\uparrow$)} [Sec~\ref{subsubsec:drate}]\\Pair-wised Topic Diversity Rating: Rates the thematic distinctiveness between two topics.}
  };
\end{scope}
\begin{scope}[yshift=4.7cm]
\node[double, boxGroup]{Intra-topic\\Semantic\\Soundness}
  child {node[children1, boxRate] {\textbf{$R_{\text{rate}}$ (rate 1-3 $\uparrow$)} [Sec~\ref{subsubsec:rrate}]\\
  Repetitiveness Rating: Rates the degree of semantic redundancy among topic words.}
  child {node[children2, boxCount] {\textbf{$R_{\text{duplicate}}$ (count $\downarrow$)} [Sec~\ref{subsubsec:rduplicate}]\\
  Duplicate Concept Detection: Identifies pairs of words that refer to the same concept.\\ 
  Adversarial Test: $AdvT_{\text{duplicate}}$ [Sec~\ref{subsubsec:advduplicate}]}}
  }
  child {node[children1, boxRate] {\textbf{$C_{\text{rate}}$ (rate 1-3 $\uparrow$)} [Sec~\ref{subsubsec:crate}]\\
  Coherence Rating: Rates the semantic unity and interpretability of a topic’s word set.}
  child {node[children2, boxCount] {\textbf{$C_{\text{outlier}}$ (count $\downarrow$)} [Sec~\ref{subsubsec:coutlier}]\\
  Outlier Detection: Flags words that are semantically inconsistent with the main theme.\\
  Adversarial Test: $AdvT_{\text{outlier}}$ [Sec~\ref{subsubsec:advoutlier}]}}
  };
\end{scope}
\begin{scope}[yshift=7.7cm]
\node[solo, boxGroup]{Lexical Validity of Topic Words}
  child {node[children1, boxRate] {\textbf{$L_{\text{rate}}$ (rate 1-3 $\uparrow$)} [Sec~\ref{subsubsec:lrate}]\\ 
  Lexical Validity Rating: Assesses if topic words are individually valid and interpretable.}
  child {node[children2, boxCount] {\textbf{$L_{\text{nonword}}$ (count $\downarrow$)} [Sec~\ref{subsubsec:lnonword}]\\ 
  Nonword Detection: Flags any malformed, garbled, or nonsensical words.
  \\Adversarial Test: $AdvT_{\text{nonword}}$ [Sec~\ref{subsubsec:advnonword}]}}
  };
\end{scope}
\end{tikzpicture}
\vspace{5mm}
\caption{Overview of our LLM-based evaluation metrics. Group by different evaluation \colorbox{groupbg}{dimensions}, there are \colorbox{ratebg}{rating-based} and \colorbox{countbg}{counting-based} metrics.}
\label{fig:llm-based-metric}
\end{figure*}
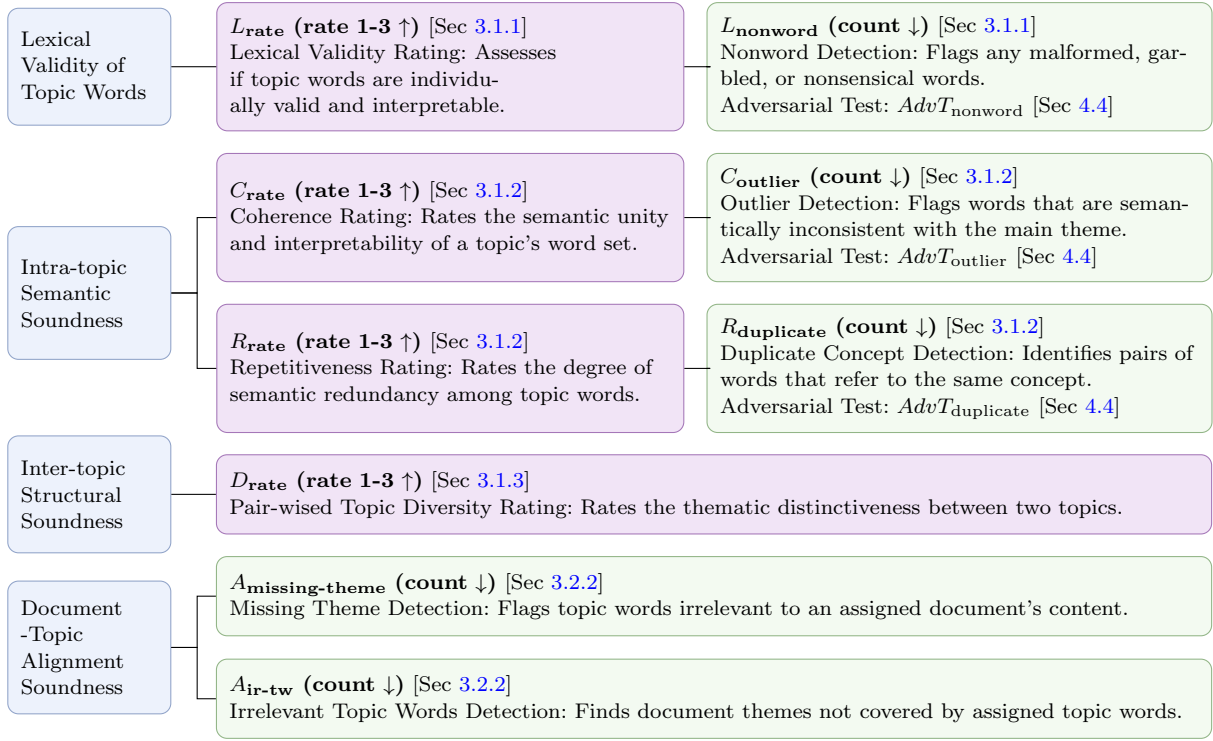

\section{Purpose-oriented Evaluation: Reconstructing Topic Quality Standards from Real Tasks}\label{sec:purpose-oriented}

Although topic modeling has been evaluated using a wide array of metrics, such as coherence, diversity, and uniqueness, these criteria are typically defined in isolation from the practical tasks that topic models are intended to support. As discussed in Section~\ref{sec:evolution}, early evaluation strategies often prioritized statistical fit or intra-topic association, while more recent work has expanded to include metrics for semantic distinctiveness and document-level alignment. However, despite these advancements, many evaluation approaches still fall short of establishing a clear connection between model quality and downstream utility.

This disconnect has produced what may be called an evaluation illusion: topic models may achieve strong performance on internal metrics, yet fall short when deployed for downstream tasks, such as exploration, classification, or retrieval. For instance, a highly coherent topic word set might consist entirely of color terms or synonyms, which contributes little to document organization or user understanding. In short, coherence does not imply practicality.

To bridge this gap, we propose a purpose-oriented evaluation framework that reverses the usual logic of assessment. Instead of evaluating topics solely on intrinsic properties, we begin with the external purposes of topic modeling, what users or systems aim to accomplish, and derive from them a set of functional quality dimensions. This perspective grounds evaluation in practical goals, enabling the design of metrics that are both task-relevant and semantically informed.

We identify three core purposes of topic modeling, each associated with specific quality requirements:
\begin{enumerate} 
    \item \textbf{Content Understanding} (in Section~\ref{subsec:PurposeI}): Topics serve as semantic summaries to help humans interpret large text collections; 
    \item \textbf{Document Labeling and Classification} (in Section~\ref{subsec:PurposeII}): Topics function as soft, machine-interpretable labels for document organization and categorization;
    \item \textbf{Topic-based Retrieval and Summarization} (in Section~\ref{subsec:PurposeIII}): Topics act as intermediaries for retrieving or distilling content based on thematic structure.
\end{enumerate}

In the sections that follow, we detail each purpose, articulate its evaluation logic, and organize existing evaluation metrics according to their corresponding dimensions. For dimensions where traditional metrics remain insufficient, we introduce LLM-based evaluation methods designed to capture quality aspects more closely aligned with real-world task performance (shown in Figure~\ref{fig:llm-based-metric}).

\subsection{Purpose I: Content Understanding — Supporting Human Interpretation of Large Corpora}\label{subsec:PurposeI}

A foundational purpose of topic modeling is to assist human users in understanding the structure, themes, and semantic landscape of large document collections. In this setting, topics are not hidden features for downstream tasks, but explicit representations to be read, interpreted, and organized by analysts, scholars, or decision-makers. They function as conceptual summaries, entry points for exploration, thematic mapping, diachronic tracking, or discourse profiling.

Applications abound across domains: topic models have been used to trace research trends in scientific corpora \cite{griffiths-2004-gibbs, lee-2021-sustainability, mejia-2021-bibliometric, dwivedi-2023-evolutionAI}, uncover the evolution of political narratives in media \cite{koltsova-2013-publicRussian, deMelo-2021-covid19Brazil}, and visualize the organization of knowledge. In each case, the utility of a topic model depends on whether its outputs are understandable and structurally meaningful to humans.

To support this interpretive purpose, we identify three key quality dimensions.

\begin{itemize} 
    \item \textit{Dimension 1}: Lexical Validity of Topic Words: Are individual topic words clear, interpretable, and well-formed?
    \item \textit{Dimension 2}: Intra-topic Semantic Soundness: Do the topic word set form a coherent theme?
    \item \textit{Dimension 3}: Inter-topic Structural Soundness: Do topics collectively form a distinct and well-organized thematic structure?
\end{itemize} 

We describe each dimension below and propose LLM-based metrics that complement and improve upon traditional evaluation strategies.

\begin{table*}[]
\promptcaption{Prompt for evaluating Lexical Validity.}
\label{prompt:lexical-validity}
\begin{tabular}{p{15cm}}
\toprule
\textbf{Prompt for Lexical Validity Rating metric} $L_{\text{rate}}$\\
Given a topic word set {[}TOPIC WORDS{]} produced by a topic model, assess whether the words are syntactically well-formed, lexically valid, and understandable in isolation. Assign an ordinal rating from 1 to 3, where 1 indicates serious issues (e.g., malformed tokens, nonwords, vague terms);
2 indicates mostly valid with minor issues;
3 indicates clean, readable, and suitable for human interpretation.
The rate is: [RATE]\\  
\midrule
\textbf{Prompt for Nonwords Detection metric} $L_{\text{nonword}}$\\
Given a topic word set {[}TOPIC WORDS{]} produced by a topic model, identify words that are garbled or malformed (e.g., typos, broken strings, random characters) or extremely rare abbreviations with unclear form or interpretation. 
Return a comma-separated list of these words or tokens or [ ] if there are none.
The invalid words or tokens are: [WORD LIST] \\ 
\midrule
\textbf{Prompt for Adversarial Test} $AdvT_{\text{nonword}}$\\
(Note: similar as prompt for $L_{\text{nonword}}$, only adding a explanation of why this is invalid lexical.)\\
\bottomrule
\end{tabular}
\end{table*}

\subsubsection{Dimension 1: Lexical Validity of Topic Words.}

Before evaluating a topic model’s coherence or structure, it is necessary to verify the lexical validity of its topic words, that is, whether they are well-formed, interpretable, and recognizable as legitimate linguistic units. This dimension ensures that topic words are usable as semantic descriptors, a prerequisite often overlooked by conventional metrics.
In practice, especially with noisy or low-resource corpora, topic models frequently produce malformed or obscure terms, nonwords, garbled strings, or ambiguous abbreviations that undermine interpretability. These errors often pass undetected by coherence metrics, which assume linguistic correctness.
We introduce two LLM-based metrics to assess lexical quality directly and independently from semantic cohesion, enabling early detection of token-level issues that hinder topic readability and downstream use.

\label{subsubsec:lrate} \textbf{Lexical Validity Rating ($L_{\text{rate}}$)} assesses the linguistic clarity of a topic word set.  
Given a topic word set $\mathcal{T} = \{w_1, w_2, \dots, w_M\}$, we prompt a LLM to return a scalar rating:
$L_{\text{rate}}(\mathcal{T}) \in \{1, 2, 3\}$
This score reflects the model’s judgment of whether the words in $\mathcal{T}$ are syntactically well-formed, lexically valid, and understandable in isolation, independent of thematic coherence. A score of 3 indicates a clean and readable topic; a score of 2 suggests minor issues or ambiguity; and a score of 1 reflects severe Lexical Validity issues such as malformed tokens, garbled words, or obscure terms. Compared to rule-based filters or dictionaries, LLMs are more robust to spelling variation, morphological irregularities, and multilingual terms.

\noindent \label{subsubsec:lnonword} \textbf{Nonword Detection ($L_{\text{nonword}}$)} identifies topic words that do not correspond to any linguistically valid or recognizable form. The detection focuses on malformed tokens, such as typos, gibberish, broken symbols, or encoding artifacts that are unlikely to carry semantic meaning.
Given a topic word set $\mathcal{T} = \{w_1, w_2, \dots, w_M\}$, we prompt the LLM to examine each word individually and determine whether it satisfies general linguistic well-formedness. A word is judged to be a “nonword” if it does not follow typical morphological, orthographic, or lexical patterns recognizable.
For notational clarity, we denote the LLM's internal decision using a boolean function $\texttt{IsNonword}(w_i)$ that returns \texttt{True} if $w_i$ is considered invalid, and \texttt{False} otherwise. The final output is the subset of topic words flagged as nonwords:
$L_{\text{nonword}}(\mathcal{T}) = \{w_i \in \mathcal{T} \mid \texttt{IsNonword}(w_i)\}$
This metric provides fine-grained diagnosis of token-level quality issues, supporting downstream tasks such as vocabulary filtering, token correction, or corpus cleaning.

These two metrics: $L_{\text{rate}}$ and $L_{\text{nonword}}$, isolate surface-level lexical problems and provide actionable signals for cleaning malformed topics before semantic evaluation or downstream use.
For prompt examples corresponding to each metric, see Prompt~\ref{prompt:lexical-validity}.

\begin{table*}[]
\promptcaption{Prompt for evaluating coherence.}
\label{prompt:coherence}
\begin{tabular}{p{15cm}}
\toprule
\textbf{Prompt for Coherence Rating metric} $C_{\text{rate}}$\\
Given a topic word set {[}TOPIC WORDS{]} produced by a topic model, assess the degree of semantic consistency among the words in the context of the topic.
Assign an ordinal rating from 1 to 3 for coherence, where 1 indicates that the words are mostly unrelated, and 3 indicates that the words are highly related and form a clear, unified theme.
The rate is: [RATE]\\  
\midrule
\textbf{Prompt for Outlier Detection metric} $C_{\text{outlier}}$\\
Given a topic word set {[}TOPIC WORDS{]} produced by a topic model, identify the words that do not semantically belong to the same conceptual theme as the others. 
Put them into a comma-separated list.
The semantically inconsistent words are: [WORD LIST]\\
\midrule
\textbf{Prompt for Adversarial Test} $AdvT_{\text{outlier}}$\\
(Note: similar as prompt for $C_{\text{outlier}}$, only adding ``If all words consistent, reply: No outliers''.)\\
\bottomrule
\end{tabular}
\end{table*}

\subsubsection{Dimension 2: Intra-topic Semantic Soundness.}

Beyond lexical validity, a well-formed topic should exhibit semantic coherence, its words should collectively form a meaningful, nameable concept. This dimension targets human-perceived coherence: the extent to which a topic evokes a unified theme interpretable by readers.

To approximate this human judgment, prior work has introduced automatic topic coherence metrics that rely on word co-occurrence or semantic similarity:

\begin{itemize} 
    \item $C_{\text{UMass}}$~\cite{mimno2011optimizing}: based on log conditional probabilities between word pairs in a reference corpus; 
    \item $C_{\text{UCI}}$~\cite{Newman2010TMforDL}: based on pairwise pointwise mutual information (PMI) within a sliding window; 
    \item $C_{\text{NPMI}}$~\cite{bouma2009NPMI, aletras2013TC}: a normalized PMI score that better reflects human judgments; 
    \item $C_{\text{V}}$~\cite{roder2015CV}: combines sliding-window co-occurrence, NPMI, and contextual graphs into a composite score;
    \item External Word Embedding Coherence~\cite{ding2018coherenceNTM}: measures cosine similarity between topic words using pretrained embeddings. 
\end{itemize}
These metrics enable scalable tuning, but fall short in sparse or domain-specific contexts, where semantically related words may rarely co-occur or behave inconsistently in embeddings. More importantly, these metrics often diverge from human judgments of interpretability.
For example, \citet{doogan-buntine-2021-TopicTwaddle} showed that the LDA-generated topic “red, wear, flag, blue, gold, black, tape, tie, green, iron” received a high $C_{\text{NPMI}}$ score (ranked 9th of 60 topics), yet human annotators were unable to assign it a meaningful label. This example underscores a key limitation of automated coherence metrics: they can overestimate quality when topics lack a coherent conceptual core.

To better approximate human judgments, we draw on two established paradigms: the coherence rating task~\cite{newman2010automaticHuman}, which asks humans to rate topics on a three-point scale based on coherence, meaningfulness, and labelability. The second is the word intrusion task~\cite{chang2009HumanTeaLeaves}, which measures whether a topic forms a strong conceptual frame by testing whether the intruding word (one out of six) can be accurately identified as the outlier. Recent work by \citet{stammbach2023revisitingLLM} demonstrated that LLMs effectively simulate these human evaluations: their coherence ratings correlate more strongly with human judgments than traditional automated metrics.
These paradigms emphasize not just lexical proximity, but whether a topic evokes a coherent and interpretable mental model, something LLMs, with their contextual reasoning abilities, are well positioned to approximate.

Building on these foundations, we propose four LLM-based metrics that reflect different facets of human-like evaluation. Two provide global assessments: coherence rating and repetitiveness rating estimate overall conceptual focus and semantic variety. The other two, outlier detection and duplicate concept detection, offer localized insight, revealing whether a low score stems from off-topic noise, or whether a high score is inflated by excessive repetition rather than genuine cohesion. Together, these metrics move beyond abstract scoring to provide interpretable signals for model refinement and topic quality control.

\begin{table*}[]
\promptcaption{Prompt for evaluating repetitiveness.}
\label{prompt:repetitiveness}
\begin{tabular}{p{15cm}}
\toprule
\textbf{Prompt for Repetitiveness Rating metric} $R_{\text{rate}}$\\
Given a topic word set {[}TOPIC WORDS{]} produced by a topic model, evaluate if there are semantically equivalent words.
Assign an ordinal rating from 1 to 3 for repetitiveness, where 1 indicates highly repetitive with significant semantic overlap, and 3 indicates minimal repetition with diverse and distinctive words. \\
The rate is: [RATE]\\
\midrule
\textbf{Prompt for Duplicate Concept Detection metric} $R_{\text{duplicate}}$\\
Given a topic word set {[}TOPIC WORDS{]} produced by a topic model, identify pairs of words that refer to the concepts or ideas that are exactly the same (not just related or similar). Provide each pair as a tuple in a comma-separated list. \\
The word pairs are: [WORD LIST] \\ 
\midrule
\textbf{Prompt for Adversarial Test} $AdvT_{\text{duplicate}}$\\
(Note: similar as prompt for $R_{\text{duplicate}}$, only adding an [ANCHOR] to be paired with the identified word.)\\
\bottomrule
\end{tabular}
\end{table*}

\label{subsubsec:crate} \textbf{Coherence Rating ($C_{\text{rate}}$)} evaluates the semantic coherence of a topic from a human perspective. Given a topic word set $\mathcal{T} = \{w_1, w_2, \dots, w_M\}$, an LLM is prompted to assess whether these words form a unified, interpretable concept. The model assigns an ordinal rating $C_{\text{rate}}(\mathcal{T}) \in \{1, 2, 3\}$, where 3 indicates a clearly cohesive and nameable topic, 2 suggests partial or vague coherence, and 1 denotes incoherence. This metric directly simulates human judgments of interpretability, offering an alternative to traditional statistical measures. It has been shown to correlate more strongly with human ratings than conventional automated coherence metrics~\cite{stammbach2023revisitingLLM}. For prompt example see Prompt~\ref{prompt:coherence}.

\label{subsubsec:coutlier} \textbf{Outlier Detection ($C_{\text{outlier}}$)} identifies topic words that semantically disrupt the internal coherence of a topic. It simulates the word intrusion task~\cite{chang2009HumanTeaLeaves}, where a human is asked to find the word that “does not belong.” Given a topic word set $\mathcal{T} = \{w_1, w_2, \dots, w_M\}$, we prompt the LLM to assess whether each word $w_i$ aligns with the dominant semantic theme implied by the rest of the words in $\mathcal{T}$.
We treat this as a boolean decision process implicitly performed by the LLM. For each $w_i$, we assume an internal predicate $\texttt{IsOutlier}(w_i, \mathcal{T})$ that returns \texttt{True} if $w_i$ is judged inconsistent with the main conceptual focus of the topic, and \texttt{False} otherwise. The final output is the set of topic words identified as outliers:
$C_{\text{outlier}}(\mathcal{T}) = \{w_i \in \mathcal{T} \mid \texttt{IsOutlier}(w_i, \mathcal{T})\}$
This metric provides word-level interpretability and supports fine-grained diagnosis of coherence failures, complementing global coherence scores such as $C_{\text{rate}}$. For prompt example see Prompt~\ref{prompt:coherence}.

\label{subsubsec:rrate} \textbf{Repetitiveness Rating ($R_{\text{rate}}$)} assesses whether a topic's coherence arises from true semantic focus or is instead artificially inflated by repetitive or synonymous terms. This metric quantifies the internal redundancy of a topic by prompting an LLM to evaluate how much overlap or semantic duplication exists among the topic words. Given a topic word set $\mathcal{T} = \{w_1, w_2, \dots, w_M\}$, the model is instructed to rate the set on a three-point scale: $R_{\text{rate}}(\mathcal{T}) \in \{1, 2, 3\}$, where 3 indicates minimal redundancy and high diversity, 2 indicates moderate repetition, and 1 signals high redundancy with limited conceptual breadth. This score complements $C_{\text{rate}}$ by distinguishing genuine coherence from superficial repetition. For prompt example see Prompt~\ref{prompt:repetitiveness}.

\label{subsubsec:rduplicate} \textbf{Duplicate Concept Detection ($R_{\text{duplicate}}$)} identifies word pairs within a topic that express the same or nearly identical concepts. Unlike $R_{\text{rate}}$, which produces an aggregate score of redundancy, this metric offers fine-grained interpretability by explicitly locating semantic duplications within the topic.
Given a topic word set $\mathcal{T} = \{w_1, w_2, \dots, w_M\}$, we prompt the LLM to evaluate every pair $(w_i, w_j)$ with $i \ne j$ and determine whether the two terms refer to the same underlying concept. This includes synonymy, paraphrasing, or highly overlapping meaning in context.
For notation, we define a boolean relation $\texttt{IsDuplicate}(w_i, w_j)$ that returns \texttt{True} if $w_i$ and $w_j$ are judged to express the same concept. The final output is the set of redundant word pairs:
$R_{\text{duplicate}}(\mathcal{T}) = \{(w_i, w_j) \mid \texttt{IsDuplicate}(w_i, w_j),\; i \ne j\}$
This metric is particularly useful for detecting cases where a topic’s apparent coherence arises from lexical repetition rather than semantic diversity. It enables precise diagnosis and supports refinement of topic quality by discouraging redundancy-based coherence. For prompt example see Prompt~\ref{prompt:repetitiveness}.

\begin{table*}[]
\promptcaption{Prompt for evaluating diversity.}
\label{prompt:diversity}
\begin{tabular}{p{15cm}}
\toprule
\textbf{Prompt for Pair-wised Topic Diversity Rating metric} $D_{\text{rate}}$\\
Given two groups of topic words: Group 1: {[}TOPIC WORDS 1{]}, Group 2 {[}TOPIC WORDS 2{]}, analyze the themes represented by the two groups.
Assign an ordinal rating from 1 to 3 based on the degree of thematic distinctiveness between the two groups:
Rate 1: Partial overlapping themes.
Rate 3: Highly distinctive themes.\\
The rate is: [RATE] \\
\bottomrule
\end{tabular}
\end{table*}

Together, these four metrics provide a multifaceted assessment of topic cohesion. A well-formed topic should receive a high coherence rating ($C_{\text{rate}}$) and low outlier detection count ($C_{\text{outlier}}$), indicating a focused and interpretable theme. At the same time, it should exhibit low redundancy, reflected in a high repetitiveness rating ($R_{\text{rate}}$) and a minimal number of semantically duplicate pairs ($R_{\text{duplicate}}$).
While the two rating metrics quantify the overall cohesion and variety of a topic, the two detection metrics offer interpretive insight into the internal structure, revealing whether weaknesses stem from off-topic intrusions or superficial repetition. This combination enables not just evaluation, but diagnosis: by pinpointing failure modes, the metrics support targeted refinement, informed pruning, and more reliable topic selection in human-facing applications.

\subsubsection{Dimension 3: Inter-topic Structural Soundness.}

Even when individual topics are coherent, a model may still be ineffective if the overall topic set is redundant, fragmented, or poorly organized. This dimension evaluates the structural soundness of the topic collection, whether the topics partition the semantic space clearly and meaningfully.
Traditional evaluation methods primarily rely on topic-to-topic similarity or diversity metrics. Token-overlap-based measures include: 
\begin{itemize} 
    \item Topic Uniqueness ($D_{\text{TU}}$)~\cite{nan2019TopicUniqueness}: the number of top-$k$ tokens in a topic that are exclusive to that topic; 
    \item Topic Redundancy ($D_{\text{TR}}$)~\cite{Burkhardt2019TopicRedundancy}: the average token overlap across all topic pairs; 
    \item Topic Diversity ($D_{\text{TD}}$)~\cite{dieng2020ETM}: the proportion of unique tokens among all top-$k$ topic words. 
\end{itemize}
Rank-based evaluation is exemplified by: 

\begin{itemize} 
    \item Inversed Rank-Biased Overlap ($D_\text{IRBO}$)~\cite{bianchi2021CombinedTM}: a rank-biased overlap metric adapted to emphasize dissimilarity between ranked token lists. 
\end{itemize}
Word embedding-based approaches include: 

\begin{itemize} 
    \item $D_\text{WE-CD}$~\cite{bianchi2021EmbedingCentroidTD}: cosine distance between centroid vectors of topic embeddings; 
    \item $D_\text{WE-PD}$~\cite{Terragni2021EmbedingTD}: average pairwise distance between all topic embeddings; 
    \item $D_\text{WE-IRBO}$~\cite{Terragni2021EmbedingTD}: rank-aware semantic dissimilarity using word embeddings. 
\end{itemize}
While these metrics provide valuable signals, each has notable limitations. Token-overlap methods overlook semantic equivalence, failing to capture near-duplicate topics with distinct surface forms. Embedding-based metrics, while semantically informed, often lack interpretability, making it difficult to identify which topics are redundant. 
To address these gaps, we propose an LLM-based evaluation metric:

\noindent \label{subsubsec:drate} \textbf{Pair-wised Topic Diversity Rating ($D_{\text{rate}}$)} quantifies the semantic distinctiveness between two topics from a human-centered perspective. Given a pair of topic word sets $\mathcal{T}_i$ and $\mathcal{T}_j$, each consisting of $k$ top-ranked words, a LLM is prompted to rate their semantic overlap on a three-point scale:
$D_{\text{rate}}(\mathcal{T}_i, \mathcal{T}_j) \in \{1, 2, 3\}$
A rating of 3 indicates that the topics are clearly distinct, 2 denotes partial semantic overlap or shared aspects, and 1 suggests substantial redundancy or conceptual duplication. This metric helps assess whether the topic set spans diverse semantic regions or exhibits excessive overlap across themes.

Applying this LLM-based evaluation across all topic pairs enables a deeper semantic characterization of the topic set: identifying which topics are distinct, which are entangled, 
These insights can inform downstream applications such as topic merging
and structured summarization.
Prompt example used to elicit LLM judgments are provided in Prompt~\ref{prompt:diversity}.

\begin{table*}[]
\promptcaption{Prompt for evaluating topic-document alignment.}
\label{prompt:document-topic-alignment}
\begin{tabular}{p{15cm}}
\toprule
\textbf{Prompt for Irrelevant Topic Words Detection metric} $A_{\text{ir-tw}}$\\
Given a document: {[}DOCUMENT{]} and a topic word set {[}TOPIC WORDS{]}, identify which topics in the word list are not relevant to the document. \\
Return these extraneous topics, or [ ] if all topics in the word list are relevant to the document. \\
Return the extraneous topics list or [ ]: [TOPIC WORDS/[ ]]\\
\midrule
\textbf{Prompt for Missing Themes Detection metric} $A_{\text{missing-theme}}$\\
Given a document: {[}DOCUMENT{]} and a topic word set {[}TOPIC WORDS{]}, identify which themes present in the document are not included in the topic word set. \\
Return these missing themes, or [ ] if all themes from the document are included in the word list. \\
Return the missed themes list or [ ]: [MISSING THEMES/[ ]]\\
\bottomrule
\end{tabular}
\end{table*}

\subsection{Purpose II: Document Labeling and Classification}\label{subsec:PurposeII}

A key application of topic modeling is to treat topics as machine-interpretable labels for documents. Unlike human-centered interpretability (Purpose I in Section~\ref{subsec:PurposeI}), the focus here is whether topic assignments support large-scale organizational tasks, such as corpus structuring, indexing, and classification. Topics act as latent tags that help organize documents and improve discoverability.

Topic-driven labeling has been applied in domains such as legal case indexing \cite{lu-2011-legalCluster, aguiar-2022-brazilianLawsuits}, scientific data annotation \cite{tuarob-2015-annotationEnvironment}, and digital library metadata enrichment \cite{newman-2007-metadataEnrichment, cain-2016-accesslibrary}. 
Topic-label alignment also supports classification systems: partially supervised models leverage weak label signals to guide topic learning~\cite{ramage-2011-partiallyLabel}, while topic-based classifiers assign categories to previously unlabeled texts~\cite{hingmire-2013-documentClassification}.
The effectiveness of these systems depends on whether document-topic associations reflect coherent, discriminative categories aligned with document semantics. We decompose the quality requirements for this purpose into two dimensions.

\begin{itemize} 
    \item \textit{Dimension 4}: Document Clusters Boundary Clarity: Do topic assignments yield distinct, non-overlapping clusters of documents?
    \item \textit{Dimension 5}: Document-topic Alignment Soundness: Are documents assigned to topics that accurately reflect their content?
\end{itemize}
We now describe each dimension and its associated evaluation metrics.

\subsubsection{Dimension 4: Document Clusters Boundary Clarity.}

This dimension asks whether topic assignments produce semantically distinct clusters of documents.
Traditional metrics include:
\begin{itemize}
  \item Top-topic clustering~\cite{nguyen-2015-latentFeature}: assign each document to its highest-probability topic, then compute cluster quality using purity and normalized mutual information (NMI)~\cite{schutze-2008-NMI};
  \item Topic vector clustering~\cite{zhao-2021-ntmOptimal}: cluster document-topic distributions and evaluate compactness and separation by computing cluster quality using purity and NMI;
  \item Silhouette score: measure cohesion vs. separation of document embeddings conditioned on topic assignment.
\end{itemize}
As these methods are sufficient for assessing structural separation, we do not introduce LLM-based evaluation for this dimension.

\subsubsection{Dimension 5: Document–topic Alignment Soundness.}

This dimension assesses whether the topics assigned to a document meaningfully represent its semantic content, that is, whether the topics reflect what the document is actually about.
Several evaluation paradigms have been proposed to assess this alignment:
\begin{itemize}
    \item Document-topic relevance rating~\cite{bhatia2017DocumentTopicEval}: human annotators rate how well a given topic word set matches a document’s content on a 0–3 scale. Averaged scores provide a model-level measure.
    
    \item Topic intrusion task~\cite{bhatia2017DocumentTopicEval}: annotators identify the “intruder” from a set of four topics, three relevant to the document, one not. Higher accuracy suggests stronger topic–document alignment.

    \item Automated topic intrusion task~\cite{bhatia2018topicIntrusionEval}: an automated version trains a neural classifier to predict intruder topics from document–topic pairs, which achieved high correlation with human judgments
    
    \item Coverage evaluation~\cite{korenvcic2021CoverageEval}: Model-generated topics are compared against a reference set of curated topics. Matching is measured either via a trained classifier or an unsupervised similarity threshold.
    
    \item LLM-based keyword agreement~\cite{yang2024llmReadingTeaLeaves}: For each document, the model provides a topic word set; a language model then generates keywords based on the same document. Similarity metrics (e.g., lexical overlap, semantic match) are used to assess alignment.
\end{itemize}
These methods capture complementary aspects of assignment soundness, from human-judged topical relevance to automatic semantic alignment. However, they also present notable limitations. Manual evaluations are labor-intensive and hard to scale. Reference-based methods rely on curated labels or external topic sets that may not generalize. Classifier-based techniques require annotated training data and introduce model-specific biases.
To address these challenges, we introduce two LLM-based evaluations that are scalable, interpretable, and domain-adaptive:

\noindent \label{subsubsec:airtw} \textbf{Irrelevant Topic Words Detection ($A_{\text{ir-tw}}$)} evaluates whether a topic assigned to a document contains words that are semantically unrelated to the document’s content. Given a document $d$ and an assigned topic represented by the word set $\mathcal{T} = \{w_1, w_2, \dots, w_M\}$, an LLM is prompted to identify the subset of topic words that do not explicitly or implicitly relate to $d$. The result is a filtered set:
$A_{\text{ir-tw}}(d, \mathcal{T}) = \{w_i \in \mathcal{T} \mid w_i \not\sim \text{content}(d)\}$
This metric captures assignment errors in which a topic contains extraneous or off-topic elements, serving as a diagnostic tool for false-positive document-topic alignment soundness.

\begin{table*}[]
\caption{Summary statistics of datasets after preprocessing.}
\centering
\begin{tabular}{l|ccc}
\toprule
\textbf{Dataset}         & \textbf{20NG}             & \textbf{AGRIS}            & \textbf{TWEETS\_NYR}  \\ 
\midrule
Train/Test      & 11,314 / 7,532   & 454,850 / 50,703 & 4,501 / 501    \\
Token (All)     & 2,799,137        & 13,405,779       & 75,168         \\
Token (Avg.)    & 149              & 27               & 15             \\
Token (Median) & 78 & 24 & 14 \\
Token (Max.) & 12,072 & 344 & 50 \\
Vocab           & 10,000           & 10,000           & 5,000          \\
Labels Counts   & 20               & 19,469           & 10             \\ 
\bottomrule
\end{tabular}
\label{tab:statistics-datasets}
\end{table*}

\noindent \label{subsubsec:amissingtheme} \textbf{Missing Theme Detection ($A_{\text{missing-theme}}$)} detects false negatives in topic assignments by identifying salient content in the document that is not covered by any of its assigned topic words. Given a document $d$ and its associated topic word sets $\mathcal{T}_1, \mathcal{T}_2, \dots, \mathcal{T}_n$, we prompt the LLM to extract important concepts from $d$ that are semantically relevant yet absent from the union of topic words. The result is a set of missing themes:
$A_{\text{missing-theme}}(d) = \{c_1, c_2, \dots \mid c_i \notin \cup_j \mathcal{T}_j\}$
This metric captures under-representation in document–topic alignment, flagging documents whose core content is not adequately reflected in the model’s topic structure. When applied across a corpus, it also reveals which topics or conceptual areas are systematically undercovered, highlighting potential blind spots in the learned topic space.

Together, these metrics capture two key types of document–topic misalignment: $A_{\text{ir-tw}}$ identifies extraneous topic content that does not reflect the document, while $A_{\text{missing-theme}}$ detects salient ideas in the document that are not covered by any assigned topic. This dual perspective enables a more precise diagnosis of alignment quality.
Examples of LLM prompts for both boundary and assignment evaluation are provided in Prompt~\ref{prompt:document-topic-alignment}.

\subsection{Purpose III: Topic-based Retrieval and Summarization}\label{subsec:PurposeIII}

Beyond interpretation and labeling, topic modeling enables semantic access to documents via retrieval and summarization. Here, topics act as dynamic interfaces, not static summaries, used to locate, compare, and distill relevant content. Task success hinges on how well topic descriptors align with document meaning.

In retrieval and recommendation, documents are embedded in a topic space and compared via similarity metrics (e.g., cosine or Jensen–Shannon divergence). This underlies systems like topic-enhanced recommenders~\cite{wang2011CTR} and hashtag suggestions~\cite{zhao-2016-hashtagRecommendation}. Retrieval quality depends on topic assignment precision and recall: vague topics yield false positives, while missing assignments lead to false negatives.

Topic models also support unsupervised extractive summarization. Some methods extract sentences near topic centroids in embedding space as summary candidates~\cite{srivastava-2022-singleSummarization}; others match document spans against extended topic word sets~\cite{nagwani-2015-mapReduceSummarizing}.  Both assume each topic forms a coherent semantic unit reflected by its words.

These applications rely on a reliable many-to-many mapping between topics and documents. Topic word sets must accurately represent their documents, and documents must, in turn, reflect their assigned topics. Misalignment, e.g., assigning a “renewable energy policy” document to the vague topic “power, system, development”, can harm ranking or produce misleading summaries.
Effective retrieval and summarization require high-quality topic–document alignment. Two complementary aspects are critical: \begin{itemize} 
    \item \textbf{Precision}: Topics should avoid irrelevant tokens that misrepresent the document (false positives); \item \textbf{Recall}: Topics should capture the salient concepts in each document (false negatives). 
\end{itemize}
We evaluate these aspects using the same LLM-based metrics introduced in Section~\ref{subsec:PurposeII}: $A_{\text{ir-tw}}$ for irrelevant topic words and $A_{\text{missing-theme}}$ for missing document content.

Together, these metrics provide diagnostic insight into alignment quality for retrieval and summarization. Unlike labeling, which emphasizes organization, this use case demands that topics serve as precise and complete semantic handles for operational access.

\section{Experimental Setup}
\subsection{Data}

We evaluate topic models across three datasets representing diverse domains and linguistic styles: online news groups, scholarly publications, and social media. These datasets present distinct challenges and enable the assessment of topic model performance across multiple functional dimensions, including topic coherence, boundary clarity, and robustness to domain and style variation.
{The statistics of three datasets are shown in Table~\ref {tab:statistics-datasets}}.

\subsubsection*{20 Newsgroups (20NG)} This is a widely used dataset ~\cite{Hinton2009UndirectedTM, miao2017GSM-GSB-RSB, Zhang2018WHAI, wu2023ECRTM}, retrieved from the ``20news-bydate.tar.gz'' version\footnote{\url{http://qwone.com/~jason/20Newsgroups}}.  There are 11,314 documents for training and 7,532 for testing. The dataset spans a broad range of topics and is frequently used to evaluate boundary clarity and topic interpretability.
Preprocessing includes lowercasing, removal of quoted replies, email signatures, and special symbols. The original vocabulary contains 54,437 tokens. After stopword removal using Gensim’s list, it is reduced to 54,147. We retain the top 10,000 most frequent words as vocabulary for modeling.
The dataset serves as a representative case of structured yet noisy user-generated text.

\subsubsection*{AGRIS} This is a domain-specific dataset of scholarly documents retrieved from the FAO’s AGRIS database\footnote{\url{https://agris.fao.org/search/en}}, consisting of 50,067 documents (45,060 for training, 5,007 for testing). Each entry contains a title and abstract, with multi-labels drawn from AGROVOC\footnote{\url{https://agrovoc.fao.org/browse/agrovoc/en/}}, a multilingual agricultural thesaurus. To support fine-grained topic evaluation, abstracts are segmented into sentences using SaT~\cite{frohmann-etal-2024-segment}, and combined with titles, yielding 454,850 training and 50,703 test entries. We exclude non-English, extremely short, or duplicate texts. Preprocessing includes lowercasing, removal of special characters, lemmatization via SpaCy, and stopword filtering using both Gensim’s list and domain-specific frequency thresholds~\cite{Gerlach2019}. The vocabulary size is reduced from 90,604 to 13,354, from which the top 10,000 most frequent words are retained for modeling.
This dataset supports fine-grained evaluation of topic fidelity in specialized academic text.

\begin{table*}[htbp]{
\centering
\caption{Hyperparameter search space and the best configurations selected for each topic model. The right three columns specify the final settings for each dataset, chosen based on the average score of five automated metrics.}
\label{tab:hyperparameter_selection}
\begin{threeparttable}
\footnotesize
\setlength{\tabcolsep}{6.8pt}
\begin{tabular}{p{1.7cm} p{2.2cm} p{2.58cm} p{0.85cm} p{0.85cm} p{0.85cm} p{0.85cm} p{0.85cm} p{0.85cm}}
\toprule
\multirow{3}{*}{\textbf{Topic Model}} & \multirow{3}{*}{\textbf{\makecell{Hyperparameter}}} & \multirow{3}{*}{\textbf{\makecell{Grid Search Space}}} & \multicolumn{6}{c}{\textbf{Best Configuration per Dataset}} \\
\cmidrule(lr){4-9}
& & & \multicolumn{2}{c}{\textbf{20NG}} & \multicolumn{2}{c}{\textbf{AGRIS}} & \multicolumn{2}{c}{\textbf{TWEETS\_NYR}} \\
\cmidrule(lr){4-9}
& & & \textbf{K=50} & \textbf{K=100} & \textbf{K=50} & \textbf{K=100} & \textbf{K=50} & \textbf{K=100} \\
\midrule
\multirow{2}{*}{LDA} & $\alpha$ & $\{0.01, 0.1, 0.5\}$ & $0.5$ & $0.5$ & $0.5$ & $0.5$ & $0.5$ & $0.01$ \\
                     & $\eta$ & $\{0.01, 0.1, 0.5\}$   & $0.01$  & $0.01$ & $0.5$  & $0.5$ & $0.1$ & $0.01$   \\
\addlinespace
\multirow{3}{*}{ProdLDA} & \verb|hidden_size| & $\{100, 200, 400\}$ & $200$ & $200$ & $200$ & $200$ & $200$ & $200$ \\
    & \verb|dropout| & $\{0.1, 0.3, 0.5\}$ & $0.1$ & $0.1$ & $0.1$ & $0.1$ & $0.1$ & $0.5$ \\
    & \verb|learning_rate| & $\{0.001, 0.002, 0.005\}$ & $0.001$ & $0.005$ & $0.005$ & $0.005$ & $0.005$ & $0.005$ \\
\addlinespace
\multirow{3}{*}{CombinedTM} & \verb|hidden_size| & $\{100, 200, 400\}$ & $200$ & $200$ & $200$ & $200$ & $200$ & $200$ \\
                     & \verb|dropout| & $\{0.1, 0.3, 0.5\}$ & $0.1$ & $0.1$ & $0.1$ & $0.1$ & $0.1$ & $0.3$ \\
                     & \verb|learning_rate| & $\{0.001, 0.002, 0.005\}$ & $0.005$ & $0.005$ & $0.005$ & $0.002$ & $0.005$ & $0.005$ \\
\addlinespace
\multirow{4}{*}{BERTopic} & \verb|n_neighbors| & 5 to 50, step size 5 & $10$ & $10$ & $10$ & $10$ & $25$ & $35$ \\
                     & \verb|n_components| & 5 to 30, step size 5 & $5$ & $5$ & $5$ & $5$ & $20$ & $25$ \\
                     & \verb|min_cluster_size| & see detail\tnote{a} & $25$ & $13$ & $680$ & $370$ & $20$ & $9$ \\
                     & \verb|diversity| & $\{0.3, 0.5, 0.7\}$ & $0.3$ & $0.3$ & $0.3$ & $0.3$ & $0.3$ & $0.5$ \\
\bottomrule
\end{tabular}
\begin{tablenotes}
    \item[a] The search space for \verb|min_cluster_size| varied by dataset due to differences in scale and document density. For 20NG and TWEETS\_NYR, it was $10$ to $25$ with step size $5$; for the much larger AGRIS dataset, it was $340$ to $400$ with step size $30$ and $650$ to $710$ with step size $30$.
\end{tablenotes}
\end{threeparttable}}
\end{table*}

\subsubsection*{TWEETS\_NYR} This dataset consists of 5,002 New Year’s resolution tweets\footnote{\url{https://www.kaggle.com/datasets/andrewmvd/new-years-resolutions}}. After removing non-English and entries with insufficient content, we split the data into 4,501 training and 501 test entries. Preprocessing includes user/retweet removal, hashtag normalization, lemmatization, and stopword filtering based on both Gensim’s list and domain-specific frequency statistics~\cite{Gerlach2019}. 
The vocabulary is reduced from 5,477 to 5,230, from which the top 5,000 most frequent words are retained for modeling. 
As a corpus of short, informal, and noisy text, this dataset serves to evaluate model robustness under minimal context.

\subsection{Topic Models}
We evaluate four representative topic modeling approaches that span probabilistic, neural, and embedding-based paradigms:

\noindent \textbf{LDA}~\cite{blei2003LDA}: a foundational probabilistic topic model that represents documents as mixtures of topics, with each topic modeled as a distribution over words. We use the Gensim implementation\footnote{\url{https://radimrehurek.com/gensim/models/ldamodel.html}}.

\noindent \textbf{ProdLDA}~\cite{srivastava2017ProdLDA-AVITM}: a neural adaptation of LDA that leverages variational autoencoders to enhance scalability and improve topic coherence. we adopt the code provided by the OCTIS toolkit\footnote{\url{https://github.com/MIND-Lab/OCTIS}}.

\noindent \textbf{CombinedTM}~\cite{bianchi2021CombinedTM}: it integrates contextual embeddings from pre-trained transformers into the LDA framework, effectively capturing semantic nuances through deep neural embeddings. We used the official implementation\footnote{\url{https://github.com/MilaNLProc/contextualized-topic-models}}.

\noindent \textbf{BERTopic}~\cite{grootendorst2022bertopic}: combines document embeddings with a class-based TF-IDF procedure to generate coherent and interpretable topics. For this study, we configured BERTopic with UMAP for dimensionality reduction and HDBSCAN for clustering, following its standard pipeline\footnote{\url{https://maartengr.github.io/BERTopic/}}.

\subsubsection*{Hyperparameter Selection}
For each model-dataset-topic combination, we conduct exhaustive hyperparameter tuning. We define a discrete search space for each model, evaluate all configurations using five automatic metrics ($C_{\text{UMass}}$, $C_{\text{NPMI}}$, $C_{\text{V}}$, $D_{\text{TU}}$, and $D_\text{IRBO}$), and select the best-performing configuration based on the average score. 
For ProdLDA and CombinedTM, the number of training epochs was fixed at 40. 
For BERTopic, the \verb|min_samples| parameter for HDBSCAN was set proportionally as 0.8 of the \verb|min_cluster_size|. 
The complete search spaces and the final selected hyperparameters for each model and dataset are detailed in Table~\ref{tab:hyperparameter_selection}.

\begin{table*}[htbp]{
\centering
\caption{Specifications of the LLMs used for evaluation. Models marked with an asterisk (*) were 4-bit quantized using Unsloth.}
\small
\label{tab:llm_specs}
\begin{tabular}{p{2.7cm} p{2.9cm} p{9cm}}
\toprule
\textbf{LLM} & \textbf{Context Window} & \textbf{Model Identifier on Hugging Face} \\ \midrule
Gemma & 128K & \texttt{\href{https://huggingface.co/google/gemma-3-4b-it}{google/gemma-3-4b-it}} \\
Gemma-large* & 128K & \texttt{\href{https://huggingface.co/unsloth/gemma-3-27b-it-qat-bnb-4bit}{unsloth/gemma-3-27b-it-qat-bnb-4bit}} \\ \addlinespace
Llama & 128K & \texttt{\href{https://huggingface.co/meta-llama/Llama-3.1-8B-Instruct}{meta-llama/Llama-3.1-8B-Instruct}} \\
Llama-large* & 128K & \texttt{\href{https://huggingface.co/unsloth/Llama-3.3-70B-Instruct-bnb-4bit}{unsloth/Llama-3.3-70B-Instruct-bnb-4bit}} \\ \addlinespace
Mistral & 128K & \texttt{\href{https://huggingface.co/mistralai/Ministral-8B-Instruct-2410}{Ministral-8B-Instruct-2410}} \\
Mistral-large* & 128K & \texttt{\href{https://huggingface.co/unsloth/Mistral-Small-3.1-24B-Instruct-2503-bnb-4bit}{unsloth/Mistral-Small-3.1-24B-Instruct-2503-bnb-4bit}} \\ \addlinespace
Qwen & 128K & \texttt{\href{https://huggingface.co/Qwen/Qwen2.5-7B-Instruct}{Qwen/Qwen2.5-7B-Instruct}} \\
Qwen-large* & 128K & \texttt{\href{https://huggingface.co/unsloth/Qwen2.5-72B-Instruct-bnb-4bit}{unsloth/Qwen2.5-72B-Instruct-bnb-4bit}} \\ \bottomrule
\end{tabular}}
\end{table*}

\subsubsection*{Training Setting}
All topic models are trained with two topic numbers, $K{=}50$ and $K{=}100$, and each configuration is executed once after hyperparameter tuning. For each trained model, we extract the top-10 most probable words (i.e. M=10) per topic to represent the topic set. For CombinedTM and BERTopic, word embeddings are initialized using the pre-trained SentenceTransformer model \texttt{all-MiniLM-L12-v2}, which is fixed throughout training to ensure consistent semantic grounding.
To reduce computational cost while maintaining evaluation reliability, we refrain from repeated training runs. Instead, all model outputs are subjected to repeated sampling during evaluation (e.g., multiple LLM completions per metric), ensuring stability and robustness of metric estimates without requiring multiple model instantiations.

\subsubsection*{Topic Models Training Efficiency and Resource Usage}
We record the training time for each model–dataset combination using their final hyperparameter configurations. All LDA runs were executed on CPU only, requiring approximately 1 minute on 20NG, 4 minutes on AGRIS, and 10 seconds on TWEETS\_NYR. ProdLDA exhibited varying demands: 58 seconds on 20NG and 13 seconds on TWEETS\_NYR using CPU, while AGRIS required GPU acceleration, reducing training time from 40 minutes (CPU) to 10 minutes (H100). CombinedTM also benefited from GPU usage on AGRIS (20 minutes), while remaining lightweight on CPU for 20NG (2 minutes) and TWEETS\_NYR (33 seconds). BERTopic was consistently fast across datasets, with GPU-based training on H100 taking 15–22 seconds for 20NG and TWEETS\_NYR, and 3 minutes for AGRIS.

\subsection{Evaluation}

\subsubsection*{LLM-based Evaluation Setup}
We adopt eight open-source instruction-tuned LLMs to assess topic quality, as detailed in Table~\ref{tab:llm_specs}. These models were chosen to represent a diverse range of modern architectures and capabilities, including state-of-the-art context window sizes. For the large-scale models, we used 4-bit quantized versions provided by Unsloth \footnote{\url{https://unsloth.ai/}} to ensure efficient computation, while the base models were sourced directly from Hugging Face\footnote{\url{https://huggingface.co}}.

All models are evaluated under identical settings: 
temperature 0.7, stochastic decoding, and 5 sampling runs per query. 
For scalar metrics (e.g., $1\textendash3$ 
rating scores), we compute the average across runs; 
For boolean judgments, 
we apply majority voting ($\geq 3$ of $5$).

\subsubsection*{Sampling Protocol and Metric Application}
All LLM-based metrics are computed using the top-10 words of each topic ($M=10$), across the full topic set for each trained model. For document-topic alignment soundness metrics, we randomly sample up to 5 documents per topic from the model output, yielding 3,005 pairs (20NG), 2,995 pairs (AGRIS),  and 3,055 pairs (TWEETS\_NYR). These samples are fed into the corresponding LLM metric prompts, enabling localized evaluation of assignment soundness. Each prompt is issued 5 times per instance to capture output variability. For the adversarial test, we randomly sampled 100 topics from four topic models applied to the three datasets.
{ The handling of document length posed no challenge in our experiments. The maximum document length across all datasets (approx. 12,000 tokens, see Table~\ref{tab:statistics-datasets}) was well within the 128K-token context window supported by our suite of LLMs (see Table~\ref{tab:llm_specs}). Consequently, all documents were processed in their entirety without truncation. This approach ensures scalability for many digital library applications, and can be readily adapted for even larger texts by leveraging models with extended context windows.}

\begin{table*}[htbp]{
\centering
\setlength{\tabcolsep}{5.8pt}
\small 
\caption{Adversarial test accuracy across datasets. Datasets are abbreviated: \textbf{N} (20NG), \textbf{A} (AGRIS), \textbf{T} (TWEETS\_NYR). Per-test and overall averages are shown in bold.}
\label{tab:adverisal}
\begin{tabular}{@{}l S[table-format=1.2] S[table-format=1.2] S[table-format=1.2] S[table-format=1.2] | S[table-format=1.2] S[table-format=1.2] S[table-format=1.2] S[table-format=1.2] | S[table-format=1.2] S[table-format=1.2] S[table-format=1.2] S[table-format=1.2] || S[table-format=1.2]@{}}
\toprule
\multirow{2}{*}{\textbf{LLM}} & \multicolumn{4}{c|}{$AdvT_{\text{nonword}}$} & \multicolumn{4}{c|}{$AdvT_{\text{outlier}}$} & \multicolumn{4}{c||}{$AdvT_{\text{duplicate}}$} & \multicolumn{1}{c}{\multirow{2}{*}{\textbf{Overall}}} \\
\cmidrule(lr){2-5} \cmidrule(lr){6-9} \cmidrule(lr){10-13}
 & {N} & {A} & {T} & {\textbf{Avg}} & {N} & {A} & {T} & {\textbf{Avg}} & {N} & {A} & {T} & {\textbf{Avg}} & \\
\midrule
Gemma         & 0.98 & 1.00 & 0.93 & {\bfseries 0.97} & 0.75 & 0.85 & 0.95 & {\bfseries 0.85} & 0.94 & 0.87 & 0.79 & {\bfseries 0.87} & {\bfseries 0.90} \\
Gemma-large   & 1.00 & 1.00 & 0.97 & {\bfseries 0.99} & 0.91 & 0.86 & 0.99 & {\bfseries 0.92} & 0.92 & 0.81 & 0.79 & {\bfseries 0.84} & {\bfseries 0.92} \\
Llama         & 0.81 & 0.83 & 0.39 & {\bfseries 0.68} & 0.86 & 0.83 & 0.96 & {\bfseries 0.88} & 0.92 & 0.85 & 0.87 & {\bfseries 0.88} & {\bfseries 0.81} \\
Llama-large   & 0.92 & 0.99 & 0.71 & {\bfseries 0.87} & 0.86 & 0.87 & 0.97 & {\bfseries 0.90} & 0.99 & 0.94 & 0.94 & {\bfseries 0.96} & {\bfseries 0.91} \\
Mistral       & 0.68 & 0.76 & 0.34 & {\bfseries 0.59} & 0.77 & 0.81 & 0.85 & {\bfseries 0.81} & 0.92 & 0.71 & 0.75 & {\bfseries 0.79} & {\bfseries 0.73} \\
Mistral-large & 0.97 & 1.00 & 0.69 & {\bfseries 0.89} & 0.84 & 0.88 & 0.98 & {\bfseries 0.90} & 0.97 & 0.89 & 0.89 & {\bfseries 0.92} & {\bfseries 0.90} \\
Qwen          & 0.85 & 0.98 & 0.69 & {\bfseries 0.84} & 0.76 & 0.82 & 0.94 & {\bfseries 0.84} & 0.93 & 0.91 & 0.89 & {\bfseries 0.91} & {\bfseries 0.86} \\
Qwen-large    & 0.95 & 0.99 & 0.75 & {\bfseries 0.90} & 0.88 & 0.88 & 0.98 & {\bfseries 0.91} & 0.98 & 0.98 & 0.89 & {\bfseries 0.95} & {\bfseries 0.92} \\
\bottomrule
\end{tabular}}
\end{table*}

\subsubsection*{LLM Evaluation Efficiency and Resource Usage}
{All experiments were conducted on a server equipped with an Intel Xeon Gold 6448Y CPU with 32 cores, 1007GB of RAM, and five NVIDIA H100 GPUs.}
Most models operate in bfloat16 precision, while large variants (Gemma-large, Mistral-large, Llama-large, Qwen-large) are quantized to 4-bit using the Unsloth framework. Inference is served via vLLM \cite{kwon2023efficient}, which dynamically batches requests to optimize memory utilization without requiring manual batch-size tuning.

Evaluation time varies substantially across metrics. Fast-running metrics, including $L_{\text{rate}}$, $L_{\text{nonword}}$, $C_{\text{rate}}$, $C_{\text{outlier}}$, $R_{\text{rate}}$, and $R_{\text{duplicate}}$, typically complete within 4 minutes. More complex metrics such as $A_{\text{ir-tw}}$ and $A_{\text{missing-theme}}$ require approximately 18 minutes, while the most expensive ones, $D_{\text{rate}}$, take around 42 minutes. 
Each adversarial test takes roughly 3 minutes per run.

In total, evaluating all LLM-based metrics once across the entire benchmark suite, three datasets, four topic models, and two topic counts, takes approximately 44.4 GPU hours. Five runs for each means approximately 222 GPU hours. However, parallelization across five H100 GPUs enables completion within two days. Throughout all experiments, memory usage remained below 80GB. Although the framework does not currently support distributed evaluation, its modular design allows efficient scaling across models, datasets, and topic configurations with minimal overhead.

We find that the large version of each LLM takes up the vast majority of the runtime, and it depends on the specific size of the large LLMs. Specifically, Qwen-large and Llama-large take about 9 $\times$ more time compared to their smaller versions, while Gemma-large and Mistral-large take about 5 $\times$ more time than Gemma and Mistral.

{To provide a transparent measure of computational cost, we analyzed the token throughput of our framework. For the entire suite of experiments, we estimate a total usage of approximately \textbf{10.3 million input tokens} and \textbf{1.0 million output tokens}. On a per-sample basis, the token cost varies significantly across metric types. The most economical are the rating-based metrics, such as $C_{\text{rate}}$ and $R_{\text{rate}}$, which consume \textbf{120-150 input tokens} for the prompt while generating a minimal output. Detection-based tasks like $C_{\text{outlier}}$ and $R_{\text{duplicate}}$ present a different profile, with shorter prompts but more substantial outputs (up to 50 tokens) to list identified items. By a significant margin, however, the most token-intensive tasks are the document-topic alignment soundness metrics ($A_{\text{ir-tw}}$ and $A_{\text{missing-theme}}$). This is directly attributable to the necessity of including document text within their prompts, which increases for a single assessment to the \textbf{200-300 input token} range. This granular cost breakdown, dominated by document-level analysis, not only clarifies the primary drivers of computational effort but also allows other researchers to make informed trade-offs, for instance by prioritizing topic-level metrics for more rapid or resource-constrained analyses.}

\subsection{Adversarial Test}
We introduce three targeted adversarial tests to validate the reliability of key automated metrics. Each test injects a minimal, controlled perturbation into a topic word set, such as a nonword, an outlier, or a duplicate concept word, and prompts the LLM to identify it. This validation protocol creates gold-standard cases for benchmarking, ensuring the related metrics are robust to realistic errors before full-scale automated deployment.

\label{subsubsec:advnonword} {$AdvT_{\text{nonword}}$ is designed to validate the reliability of our automated nonword detection metric. It is crucial to clarify that this test serves as a validation protocol to create a gold standard test set, and is not a component of the scalable, automated application of our framework. For each test case, we begin with a clean topic word set and manually inject one lexically invalid token to simulate minimal but realistic noise. Inserted tokens fall into one of three categories: (1) misspelled or garbled strings (e.g., ``envrnmnt'', ``x\%ag8''), (2) ambiguous or obscure abbreviations, or (3) nonwords produced via character substitution. The remaining topic words are kept unchanged to preserve semantic plausibility. The LLM is then prompted to identify which tokens do not correspond to recognizable lexical forms, relying on general linguistic plausibility. This validation ensures we can trust the results of the $L_{\text{nonword}}$ metric, which is otherwise applied in a fully automated fashion during its application phase. For prompt example see Prompt~\ref{prompt:lexical-validity}.}

\label{subsubsec:advoutlier} { $AdvT_{\text{outlier}}$ is designed to assess the reliability of the outlier detection metric. Following the validation principle established for our framework, we conduct this test by simulating controlled perturbations. Inspired by the word intrusion task~\cite{chang2009HumanTeaLeaves}, for each test sample, we manually insert a semantically unrelated term (e.g., inserting \textit{``Shakespeare''} into a topic on renewable energy) into a coherent topic word list. Each modified topic contains exactly one intruding word. The LLM is then prompted to identify the outlier, allowing us to precisely evaluate its sensitivity to minimal disruptions in topical cohesion and confirm the reliability of the automated $C_{\text{outlier}}$ metric.}
Each modified topic word set contains exactly one intruding word, while the rest remain unaltered to preserve coherence. The LLM is prompted to identify the outlier based on semantic inconsistency with the remaining words. This setup enables precise evaluation of the LLM’s sensitivity to minimal disruptions in topical cohesion. For prompt example see Prompt~\ref{prompt:coherence}.

\label{subsubsec:advduplicate} {$AdvT_{\text{duplicate}}$ is designed to validate the robustness of the duplicate concept detection metric. Consistent with our validation methodology, this test relies on creating controlled test cases with known redundancies. We manually inject a word into the topic word set that expresses the same concept as an existing word (the anchor), such as a synonym or an abbreviation (e.g., “chrismas” and “xmas”). The LLM is then prompted to detect the created word pair. This process allows us to verify the reliability of the $R_{\text{duplicate}}$ metric before its full automated deployment in our framework.}
For each test case, the LLM is prompted to detect semantically redundant word pairs. The test is successful if the inserted duplicate is paired with its intended anchor. While additional pairs may be flagged, only the detection of the inserted duplicate is scored. 
For prompt example see Prompt~\ref{prompt:repetitiveness}.

\section{Results and Discussion}

\subsection{Adversarial Test Robustness Analysis}
To assess the diagnostic reliability of LLM-based metrics under deliberately corrupted inputs, we conduct three adversarial tests, $AdvT_{\text{nonword}}$, $AdvT_{\text{outlier}}$, and $AdvT_{\text{duplicate}}$, across all datasets and LLMs. Detection accuracy was evaluated over 100 adversarially modified topics per model–dataset combination,
{ with the results summarized in Table~\ref{tab:adverisal}.}

\noindent \textbf{$AdvT_{\text{nonword}}$:}
from the aspect of datasets, detection accuracy is highest on AGRIS (0.94), followed by 20NG (0.90), and lowest on TWEETS\_NYR (0.68). This reflects the challenge of distinguishing invalid tokens in informal, low-context texts like tweets, where lexical anomalies may blend into stylistic noise.
From the LLMs as evaluator's aspect, Larger models such as Gemma-large (0.99), Qwen-large (0.90), and Mistral-large (0.89) perform best. However, Gemma (0.97), despite being a small model, shows strong performance, suggesting good small-model substitutability in this task. In contrast, Mistral (0.59) and Llama (0.68) lag behind, indicating limited robustness and poor cost-performance tradeoff in lexical validation.

\begin{figure}
    \centering
    \includegraphics[width=1\linewidth]{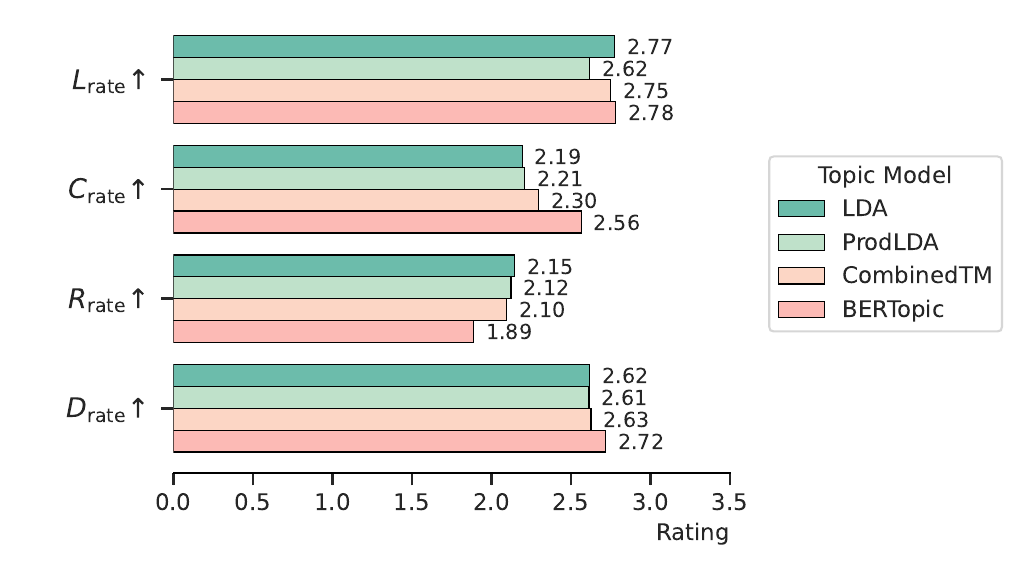}
    \caption{Result of LLM-based Metrics (rating).}
    \label{fig:compare-topic-model-rate}
\end{figure}

\noindent \textbf{$AdvT_{\text{outlier}}$:}
from the aspect of datasets, all models perform well on TWEETS\_NYR (0.95), possibly due to its sharper topical focus and less ambiguity per topic word set, while 20NG (0.83) and AGRIS (0.85) remain slightly lower.
From the LLMs as evaluator's aspect, strong performance is observed from Gemma-large (0.92), Qwen-large (0.91), and Llama-large (0.90). Notably, Llama (0.88), a smaller model, achieves performance close to its larger counterpart, indicating potential substitutability. Mistral (0.81) is the lowest, suggesting less semantic sensitivity in detecting thematic intrusions.

\noindent \textbf{$AdvT_{\text{duplicate}}$:}
from the aspect of datasets, 20NG (0.95) yields the highest accuracy, followed by AGRIS (0.87) and TWEETS\_NYR (0.85). Since topic length is fixed (top-10 words), this likely reflects the influence of domain redundancy, repetitive terminology is easier to detect in news than in scientific domains and sparse, informal tweet language.
From the LLMs as evaluator's aspect, best results come from Llama-large (0.94) and Qwen-large (0.94). However, Qwen (0.90) and Gemma (0.83), both small models, perform surprisingly well, making them efficient alternatives. In contrast, Mistral (0.73) is again the weakest, underperforming even when compared to peer small models.

\subsection{Performance of Topic Models and Dataset }

\begin{figure}
    \centering
    \includegraphics[width=1\linewidth]{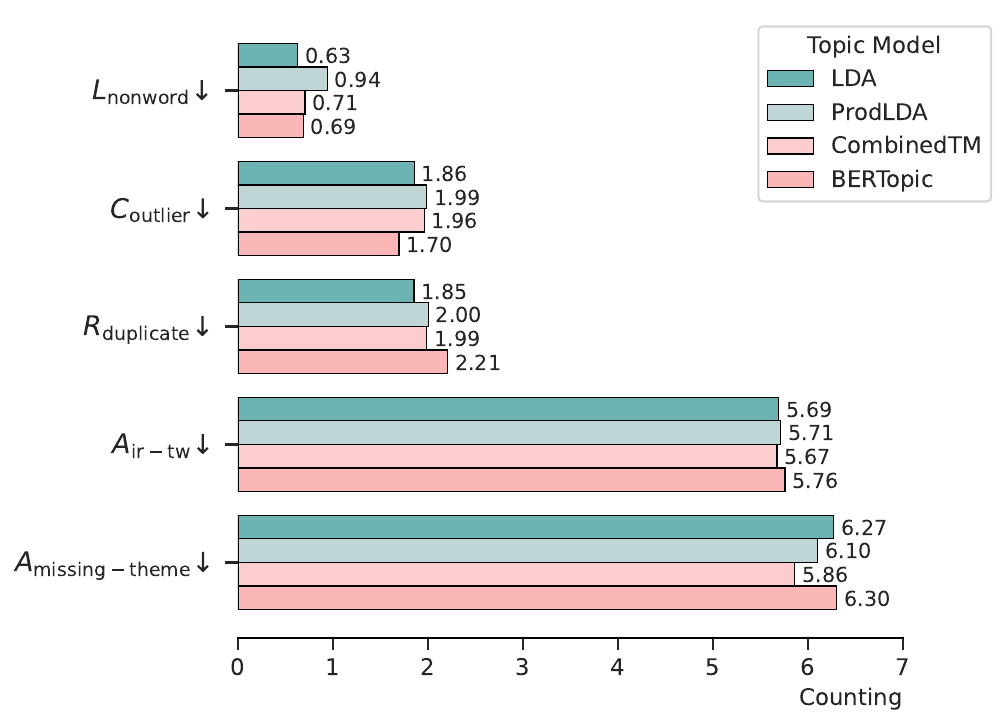}
    \caption{Result of LLM-based Metrics (counting).}
    \label{fig:compare-topic-model-count}
\end{figure}

To assess how different topic modeling methods perform under various quality metrics, we analyze results from two complementary perspectives: (1) comparison across models for fixed datasets, and (2) comparison across datasets for fixed models. Each perspective illuminates distinct aspects of model behavior, robustness, and generalization.  

\subsubsection*{Comparison Across Topic Models}  
We compare four topic models (LDA, ProdLDA, CombinedTM, and BERTopic) across nine LLM-based evaluation metrics, grouped into two categories: scalar rating metrics (higher rate is better, see Figure~\ref{fig:compare-topic-model-rate}) and error detection metrics (lower count is better, see Figure~\ref{fig:compare-topic-model-count}). Each value reflects the average across three datasets and two topic counts ($K=50$ and $K=100$).
Complete data can be found in the Supplemental Material.

\noindent \textbf{Lexical Validity.} CombinedTM and BERTopic achieve the highest scores on lexical validity ($L_{\text{rate}}$), indicating strong performance in producing fluent and interpretable topic words. LDA is competitive, while ProdLDA yields slightly lower fluency. However, in terms of nonword detection ($L_{\text{nonword}}$), which counts malformed or invalid terms, LDA achieves the lowest error rate, followed by CombinedTM. ProdLDA performs the worst, suggesting that its generated topics contain more irregular lexical entries.

\noindent \textbf{Intra-topic Semantic Soundness.} 
BERTopic achieves the highest coherence scores ($C_{\text{rate}}$). It also shows the fewest detected outliers ($C_{\text{outlier}}$), reinforcing its semantic compactness. CombinedTM follows closely, while LDA and ProdLDA score consistently lower.
In contrast to its strong coherence, BERTopic performs worst in both repetitiveness rating ($R_{\text{rate}}$) and duplicate concept detection ($R_{\text{duplicate}}$), indicating a tendency to generate thematically redundant topics. Meaning, BERTopic’s high coherence does come at the cost of excessive lexical repetition. LDA, ProdLDA, and CombinedTM all show lower redundancy levels, with LDA being particularly strong on minimizing concept duplication. 
{For instance, on the TWEETS\_NYR dataset, BERTopic produced a highly coherent topic on religion (``god, pray, jesus...'') but included the repetitive pair (``pray'', ``prayer''), a pattern we analyze further in Section~\ref{subsec:qualitative_analysis} (see Table~\ref{tab:core-examples}).}

\noindent \textbf{Inter-topic Structural Soundness.} All models perform similarly in diversity rating ($D_{\text{rate}}$), though BERTopic has a slight edge. That said, the overall gap between models on this metric is small. 

\noindent \textbf{Document-topic Alignment Soundness.} CombinedTM achieves the best performance in missing theme detection ($A_{\text{missing-theme}}$), followed by ProdLDA and LDA. BERTopic shows the highest number of missing themes, implying that its topics may fail to comprehensively reflect document content. Irrelevant topic word detection ($A_{\text{ir-tw}}$) is similarly highest for BERTopic, indicating a greater inclusion of off-topic terms.  In sum, BERTopic provides highly coherent and readable topics but at the cost of increased redundancy and alignment errors. CombinedTM offers a more balanced trade-off across metrics. LDA is surprisingly competitive, particularly in lexical precision and low redundancy. ProdLDA underperforms on most axes, likely due to instability in variational inference.

\subsubsection*{Comparison Across Datasets}  
We next examine model behavior across the three datasets, 20NG (news group style), AGRIS (scientific), and TWEETS\_NYR (short, informal text), to evaluate the impact of domain characteristics on metric outcomes.

\noindent \textbf{Lexical Validity.} AGRIS supports the best lexical validity across all models, achieving the highest $L_{\text{rate}}$ and the lowest $L_{\text{nonword}}$ values. This suggests that technical corpora with standardized vocabulary allow topic models to generate more fluent outputs. Conversely, TWEETS\_NYR presents the greatest challenge due to noise, informality, and brevity.

\noindent \textbf{Intra-topic Semantic Soundness.} 
Coherence is highest on AGRIS, moderate on 20NG, and lowest on TWEETS\_NYR. These results reflect the influence of domain regularity on internal topic consistency. The $C_{\text{outlier}}$ metric further highlights that TWEETS\_NYR topics contain more semantically incoherent terms, despite appearing reasonable in aggregate scores.
TWEETS\_NYR shows the highest ratings in $R_{\text{rate}}$, suggesting it suffers least from within-topic repetition, likely due to its shorter documents. However, $R_{\text{duplicate}}$ is lowest in AGRIS, indicating better inter-topic differentiation in the scientific dataset.

\noindent \textbf{Inter-topic Structural Soundness.} The diversity metric ($D_{\text{rate}}$) remains relatively stable across datasets, though 20NG shows slightly higher scores, possibly due to its broader topic coverage. The variation across datasets is modest.  

\noindent \textbf{Document-topic Alignment Soundness.} The most off-topic words ($A_{\text{ir-tw}}$) are detected in 20NG, followed by TWEETS\_NYR, with AGRIS performing best. Missing themes ($A_{\text{missing-theme}}$) are most frequent in 20NG and TWEETS\_NYR, both of which contain noisier or less structured content.  Overall, AGRIS emerges as the most favorable setting for generating accurate and coherent topics. 

\subsection{Correlation of Metrics}
\begin{figure*}
    \centering
    \includegraphics[width=0.99\linewidth]{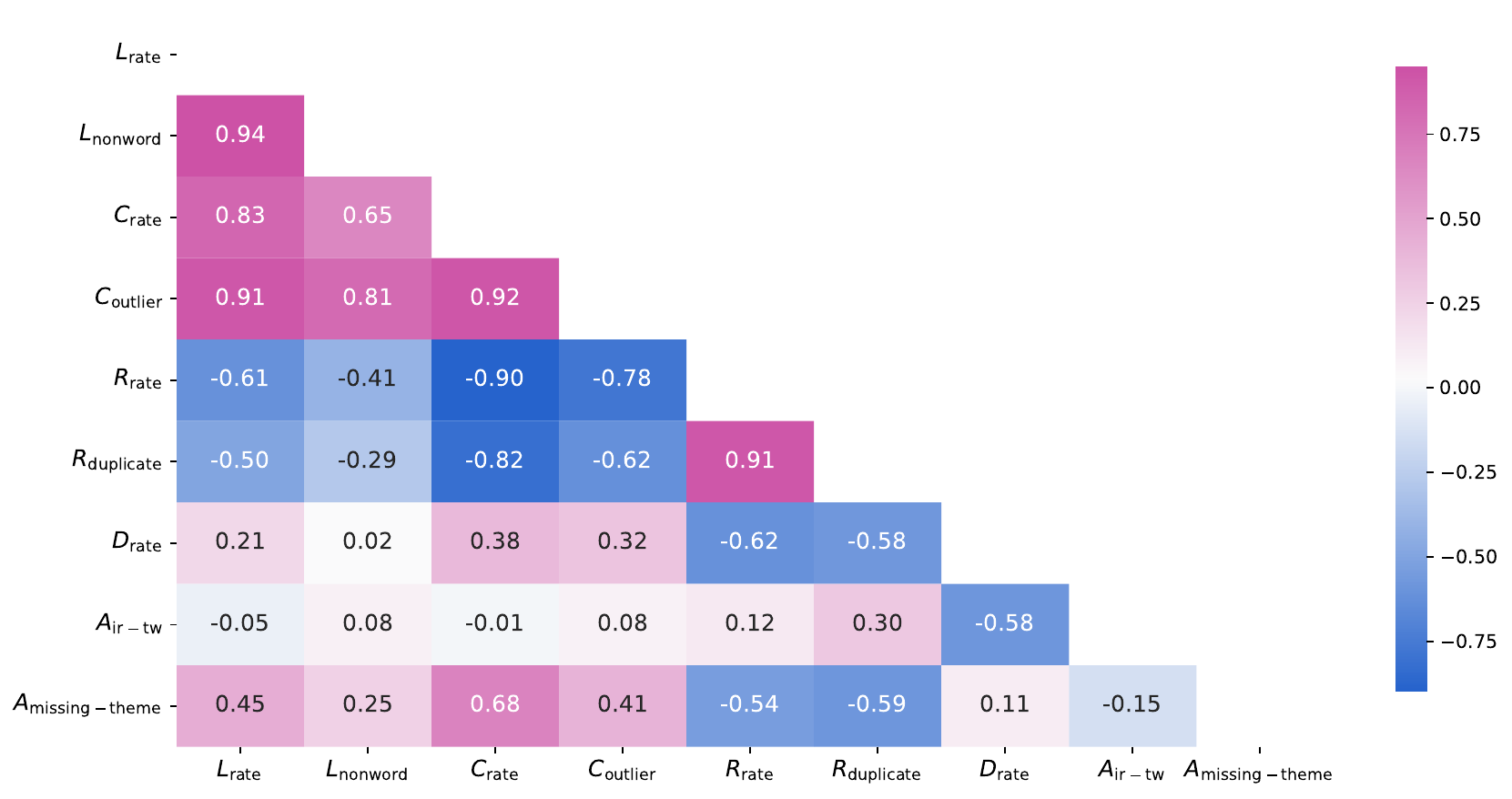}
    \caption{Correlation Heatmap of LLM-based Metrics}
    \label{fig:correlation-llm-metric}
\end{figure*}

\subsubsection*{Correlations among LLM-based Metrics}
To ensure comparability across metrics with different directional interpretations, we preprocessed all LLM-based scores before computing correlations using Pearson correlation coefficient~\cite{cohen-2009-pearson}. Specifically, we standardized all metrics such that higher values uniformly indicate better topic quality. This involved inverting the values of negatively-oriented metrics, including Nonword Detection ($L_{\text{nonword}}$), Outlier Detection ($C_{\text{outlier}}$), Duplicate Concept Detection ($R_{\text{duplicate}}$), Irrelevant Topic Word Detection ($A_{\text{ir-tw}}$), and Missing Theme Detection ($A_{\text{missing-theme}}$), by subtracting them from their maximum observed value. Correlations were then calculated using these aligned scores.

Based on the correlation heatmap~\ref{fig:correlation-llm-metric} of LLM-based evaluation metrics (after inverting the direction of negatively-oriented metrics for interpretability), we observe several consistent patterns that both validate the design of our metrics and reveal latent interactions across quality dimensions.  First, metrics designed as diagnostic companions exhibit strong internal consistency. $L_{\text{rate}}$ is highly positively correlated with $L_{\text{nonword}}$ at $r = 0.94$, indicating that when topic words are rated as lexically valid, they also tend to contain fewer malformed or meaningless tokens. Similarly, $C_{\text{rate}}$ and $C_{\text{outlier}}$ are strongly correlated ($r = 0.92$), affirming the explanatory value of $C_{\text{outlier}}$ in pinpointing incoherent elements within topics judged as globally less coherent. A parallel trend holds between $R_{\text{rate}}$ and $R_{\text{duplicate}}$, with a correlation of $r = 0.91$.  Second, an antagonistic relationship emerges between coherence and repetitiveness dimensions. $C_{\text{rate}}$ is negatively correlated with both $R_{\text{rate}}$ ($r = -0.90$) and $R_{\text{duplicate}}$ ($r = -0.82$), suggesting that highly coherent topics often avoid redundancy, while repetitive topics tend to appear semantically diffuse. These findings reflect a trade-off between internal consistency and conceptual distinctness. Similar trends appear between $C_{\text{outlier}}$ and repetitiveness metrics ($R_{\text{rate}}$ $r = -0.78$ and $R_{\text{duplicate}}$ $r = -0.62$), further supporting this observation. 

We next examine how LLM-based metrics relate to traditional automated metrics such as topic coherence and diversity, thereby assessing whether our proposed scores capture complementary or redundant aspects of topic quality.

\begin{table}[ht]
\centering
\caption{Pearson Correlation between Automated Coherence Metrics and LLM-based Metrics.}
\label{tab:correlation-llm-automated}
\begin{tabular}{lccc}
\toprule
\textbf{LLM-based Metric} & \textbf{$C_{\text{UMASS}}$} & \textbf{$C_{\text{NPMI}}$} & \textbf{$C_{\text{V}}$} \\
\midrule
$L_{\text{rate}}$        & \textbf{0.87} & 0.53 & 0.41 \\
$L_{\text{nonword}}$     & \textbf{0.70} & 0.32 & 0.22 \\
\midrule
$C_{\text{rate}}$        & \textbf{0.88} & 0.57 & 0.46 \\
$C_{\text{outlier}}$     & \textbf{0.84} & 0.44 & 0.32 \\
\midrule
$R_{\text{rate}}$        & -0.66 & -0.51 & -0.45 \\
$R_{\text{duplicate}}$   & -0.62 & -0.66 & -0.63 \\
\bottomrule
\end{tabular}
\end{table}

\begin{table*}[ht]
\centering
\caption{Comparative Average Performance of LLMs Across Tasks.}
\label{tab:llm_performance}
\begin{tabular}{@{} lcccccccc @{}}
\toprule
\textbf{Metric} & \textbf{Gemma} & \shortstack{\textbf{Gemma-}\\\textbf{large}} & \textbf{Llama} & \shortstack{\textbf{Llama-}\\\textbf{large}} & \textbf{Mistral} & \shortstack{\textbf{Mistral-}\\\textbf{large}} & \textbf{Qwen} & \shortstack{\textbf{Qwen-}\\\textbf{large}} \\
\midrule
$L_\text{rate}$ (↑) & 2.778 & 2.769 & 2.299 & 2.838 & 2.797 & 2.737 & 2.766 & 2.857 \\
$L_\text{nonword}$ (↓) & 0.258 & 0.208 & 1.220 & 1.151 & 0.166 & 0.716 & 1.677 & 0.539 \\
\midrule
$C_\text{rate}$ (↑) & 2.259 & 2.328 & 2.603 & 2.423 & 2.556 & 2.336 & 2.003 & 2.019 \\
$C_\text{outlier}$ (↓) & 1.689 & 1.707 & 1.668 & 2.189 & 1.010 & 1.631 & 2.214 & 2.895 \\
\midrule
$R_\text{rate}$ (↑) & 1.994 & 1.888 & 1.719 & 1.759 & 2.036 & 2.074 & 2.492 & 2.543 \\
$R_\text{duplicate}$ (↓) & 2.398 & 1.336 & 2.649 & 1.313 & 3.388 & 1.790 & 0.647 & 2.575 \\
\midrule
$D_{\text{rate}}$ (↑) & 2.078 & 2.929 & 2.778 & 2.908 & 2.399 & 2.105 & 2.996 & 2.957 \\
\midrule
$A_{\text{ir-tw}}$ (↓) & 4.959 & 4.396 & 5.697 & 6.020 & 5.330 & 5.278 & 6.148 & 7.853 \\
$A_{\text{missing-theme}}$ (↓) & 5.708 & 6.413 & 5.269 & 6.364 & 3.825 & 6.825 & 7.258 & 7.419 \\
\bottomrule
\end{tabular}
\end{table*}

\subsubsection*{LLM-based vs. Automated Metric Correlations}

To examine whether LLM-based evaluations are aligned with traditional automated metrics, we compute Pearson correlations between nine LLM-based metrics and five common automated scores: $C_{\text{UMASS}}$, $C_{\text{NPMI}}$, $C_{\text{V}}$, $D_{\text{TU}}$, and $D_{\text{IRBO}}$. To ensure consistency in interpretation, all LLM metrics where lower values indicate better quality (e.g., error counts) are inverted prior to correlation analysis. 
{The results are shown in Table~\ref{tab:correlation-llm-automated}.}

\noindent \textbf{Alignment with Topic Coherence Metrics.} Strong positive correlations are observed between lexical or coherence-related LLM-based metrics and traditional coherence scores. Notably, $C_{\text{UMASS}}$ aligns well with $L_{\text{rate}}$ ($r = 0.87$), $C_{\text{rate}}$ ($r = 0.88$), and $C_{\text{outlier}}$ ($r = 0.84$), confirming that LLMs and automated measures agree on identifying lexically and semantically coherent topics. $C_{\text{V}}$ also shows moderate correlations with $L_{\text{rate}}$ ($r = 0.41$), $C_{\text{rate}}$ ($r = 0.46$), suggesting that automated topic coherence scores share a partially overlapping notion of topic quality with LLM judgments.

\noindent \textbf{Tension with Repetition-based Metrics.} Metrics detecting repetition, such as $R_{\text{rate}}$ and $R_{\text{duplicate}}$, are negatively correlated with all coherence scores, particularly $C_{\text{V}}$ ($r = -0.45$ and $r = -0.63$ respectively). This supports the earlier observation that coherence metrics may inflated by topic redundancy. 

\noindent \textbf{Limited Alignment with Diversity Metrics.} Surprisingly, $D_{\text{TU}}$ and $D_{\text{IRBO}}$ do not correlate strongly with $D_{\text{rate}}$ (LLM-rated diversity). While $D_{\text{rate}}$ and $D_{\text{IRBO}}$ exhibit a modest correlation of $r = 0.31$, the overall agreement remains limited. This suggests that LLMs may be detecting aspects of thematic spread or distinctiveness not fully captured by word-overlap-based diversity scores.  

Overall, these results validate the complementarity of LLM-based metrics. While traditional metrics excel at aggregate-level signals, LLMs provide finer-grained, interpretable feedback, particularly for identifying redundancy and outliers that are otherwise overlooked.

\subsection{Comparative Analysis of LLM Labeling Performance.}

\label{subsec:llm-comparison}

To systematically compare the behavioral patterns of different LLMs as evaluators, we analyzed their average performance across all metrics and datasets, with the aggregated results presented in Table~\ref{tab:llm_performance}. This comparative analysis reveals distinct evaluator profiles for each LLM family, highlighting systematic differences in their scoring strictness and detection sensitivity.

\begin{table*}[htbp]
\centering
\caption{Behavioral Correlation Between Base and Large LLM Variants.}
\label{tab:correlation-analysis}
\begin{tabular}{llc}
\toprule
\textbf{Metric} & \textbf{High Correlation Pairs} & \textbf{Low/Inverse Correlation Pairs} \\ 
\midrule
$L_\text{rate}$ & Gemma, Mistral, Qwen ($\rho\geq0.92$) & Llama ($\rho=0.72$) \\
$L_\text{nonword}$ & Llama, Mistral, Qwen ($\rho\geq0.85$) & Gemma ($\rho=0.06$) \\ \midrule
$C_\text{rate}$ & Llama, Mistral, Qwen ($\rho\geq0.97$) & Gemma ($\rho=0.54$) \\
$C_\text{outlier}$ & Gemma, Mistral, Qwen ($\rho\geq0.77$) & Llama ($\rho=0.51$) \\ \midrule
$R_\text{rate}$ & Qwen ($\rho=0.98$), Llama ($\rho=0.84$) & Gemma ($\rho=0.44$), Mistral ($\rho=0.19$) \\
$R_\text{duplicate}$ & Gemma ($\rho=0.86$), Llama ($\rho=0.75$) & Mistral ($\rho=-0.03$), Qwen ($\rho=-0.62$) \\ \midrule
$D_\text{rate}$ & Mistral ($\rho=0.92$) & Gemma, Llama, Qwen ($\rho\leq0.60$)\\ \midrule
$A_\text{ir-tw}$ & Mistral ($\rho=0.99$), Llama ($\rho=0.99$) & Gemma ($\rho=-0.20$), Qwen ($\rho=0.31$) \\
$A_\text{missing-theme}$ & Gemma, Llama, Qwen ($\rho\geq0.78$) & Mistral ($\rho=0.72$) \\ 
\bottomrule
\end{tabular}
\end{table*}

\subsubsection*{Lexical Validity}

\noindent \textbf{$L_\text{rate}$}: Llama exhibits distinct scoring patterns compared to other LLMs (including Llama-large), with lower average scores. This behavioral difference is corroborated by its higher nonword detection rates in adversarial testing.
    
\noindent \textbf{$L_\text{nonword}$}: Both Llama variants demonstrate increased nonword extraction tendencies. Qwen shows the strongest nonword extraction behavior (mean: 1.677 vs. 0.539 for Qwen-large), though Qwen-large exhibits marginally reduced nonword rates.
{This tendency suggests that Llama models may have a lower tolerance for informal or ambiguous terms. A case in point from our qualitative analysis is the misclassification of the slang term ``mida'' as a nonword (Section~\ref{subsec:qualitative_analysis}). While other models might correctly interpret it from context, Llama's behavior indicates a stronger reliance on formal lexical validity, leading to its distinctive scoring profile in both ratings and detection tasks.}

\subsubsection*{Intra-topic Semantic Soundness}

\noindent \textbf{$C_\text{rate}$}: The Qwen family demonstrates stricter scoring behavior (mean: 2.0 vs. 2.4 for other LLMs), suggesting more conservative coherence judgments.

\noindent \textbf{$C_\text{outlier}$}: Qwen models exhibit increased outlier extraction tendencies (2.5 vs. 1.6 average outliers), aligning with their strict coherence scoring pattern. 
{A clear example from Table~\ref{table:example_outlier_detection} can be seen when evaluating the topic on voice modems: while most LLMs identified only ``clipper'' as the outlier, Qwen additionally flagged the term ``cheap'', reflecting a stricter judgment on what constitutes a relevant semantic attribute.}

\noindent \textbf{$R_\text{rate}$}: The Qwen family shows inverse behavioral patterns to $C_\text{rate}$, with higher scores indicating stricter avoidance of repetitive labels (2.5 vs. 1.9 average).
    
\noindent \textbf{$R_\text{duplicate}$}: Larger model variants (Gemma-large, Llama-large, Mistral-large) demonstrate increased duplicate extraction tendencies compared to base versions (+23\% average duplicates), suggesting model scaling impacts duplication behavior.

\subsubsection*{Inter-topic Structural Soundness}

\noindent \textbf{$D_\text{rate}$}: Significant behavioral divergence exists between model variants: Gemma-large shows 45\% higher diversity scores than Gemma. 
Mistral family demonstrates conservative scoring patterns (mean: 2.3 vs. 2.8 for other families).

\subsubsection*{Document-topic Alignment Soundness}

\noindent \textbf{$A_\text{ir-tw}$}: Qwen models demonstrate stricter association judgments, extracting 32\% more irrelevant terms than other LLMs (6.9 vs. 5.2 average).

\noindent \textbf{$A_\text{missing-theme}$}: This pattern persists in theme identification (7.34 vs. 5.73). Mistral-large shows substantial degradation vs. base Mistral (6.82 vs. 3.82).

\subsection{Resource-efficient Model Selection Through Behavioral Correlation Analysis}
\label{subsec:model-selection}

Our correlation analysis reveals scenarios where smaller base models exhibit strong behavioral alignment with their larger counterparts, enabling potential substitution for faster iteration.

Key substitution recommendations emerge from Table~\ref{tab:correlation-analysis}:
For lexical validity ($L_\text{rate}$) and coherence ($C_\text{rate}$) tasks, all base models show strong alignment ($\rho\geq0.72$) with their large variants, making them reliable proxies.
Complete correlation data can be found in the Supplementary Material.

\subsubsection*{Family-specific Substitution}
Use Mistral-base for diversity ($D_\text{rate}$) and document assignment ($A_\text{ir-tw}$) ($\rho\geq0.86$).
Employ Qwen-base for repetitiveness screening ($R_\text{rate}$: $\rho=0.98$).
Avoid Gemma-base for nonword detection ($L_\text{nonword}$: $\rho=0.10$).

\begin{table*}[htbp]
\caption{Examples of outlier detection metric $C_{\text{outlier}}$ from 20NG.}
\label{table:example_outlier_detection}
\centering
\begin{tabular}{l|p{4.2cm}|p{4.2cm}|p{4.2cm}}
\toprule
\textbf{LLM} & \textbf{Example 1} & \textbf{Example 2} & \textbf{Example 3} \\
\cmidrule{2-4}
& \textit{monitor, screen, centris, adapter, mac, viewsonic, resolution, spec, vram, cable}
& \textit{voice, modem, dsp, compression, cheap, clipper, software, phone, hardware, data}
& \textit{faq, soda, posting, berkeley, monthly, info, archive, associate, revision, document} \\
\midrule
Gemma   & \textit{mac, spec}       & \textit{clipper} & \textit{soda, berkeley} \\
Llama   & \textit{mac}             & \textit{clipper} & \textit{soda, monthly} \\
Mistral & \textit{mac, centris}    & \textit{clipper} & \textit{soda} \\
Qwen    & \textit{vram, mac, spec} & \textit{clipper, cheap}  & \textit{soda, monthly, berkeley} \\
\bottomrule
\end{tabular}
\end{table*}

\subsubsection*{Risk Scenarios}
Mistral/Qwen-large exhibit inverse duplication patterns vs. their base models ($R_\text{duplicate}$: $\rho<0$).
Llama-base shows poor outlier detection alignment ($C_\text{outlier}$: $\rho=0.18$).

\subsection{Qualitative Analysis\label{subsec:qualitative_analysis}}
In this section, we provide a qualitative analysis of representative examples to explore discrepancies and patterns in nonword detection, outlier detection, duplicate concept detection, and BERTopic coherence-repetitiveness trade-off.

\subsubsection*{Nonword Misclassification}
Gemma misclassifies ``mida'' (TWEETS\_NYR ID 44, [meet fight fangirl deepthroat xmas remote imply physically folder colleen mida]), a valid slang term, as a nonword, likely due to its reliance on strict tokenization rules rather than contextual understanding. This reflects a precision-recall tradeoff: smaller models like Gemma prioritize flagging structural anomalies (e.g., symbols/numbers) but struggle with dialectal or informal terms, leading to false negatives. Such errors highlight challenges in balancing rule-based heuristics with semantic adaptability for robust lexical validation tasks.

\begin{table*}[htbp]
\caption{Examples of duplicate concept detection metric $R_{\text{duplicate}}$ from 20NG.}
\label{table:example_repetitiveness}
\centering
\begin{tabular}{l|p{6.7cm}|p{6.7cm}}
\toprule
\textbf{LLM} & \textbf{Example 1} & \textbf{Example 2} \\
\cmidrule{2-3}
& \textit{faq, soda, posting, berkeley, monthly, info, archive, associate, revision, document} 
& \textit{printer, deskjet, ink, laserwriter, toner, printing, laserjet, inkjet, adobe, epson} \\
\midrule
Gemma
& \textit{(document, archive)}
& \textit{(printer, printing), (deskjet, inkjet), (laserwriter, laserjet)}\\
\midrule
Llama
& \textit{(document, archive), (faq, info), (revision, document)}
& \textit{(inkjet, ink), (laserjet, laserwriter)} \\
\midrule
Mistral
& \textit{(archive, revision), (faq, info)} 
& \textit{(deskjet, inkjet), (printer, deskjet), (laserwriter, laserjet)} \\
\midrule
Qwen
& \textit{(faq, info), (document, archive)}
& \textit{(deskjet, laserjet), (laserwriter, laserjet), (deskjet, inkjet), (laserwriter, inkjet), (epson, adobe)} \\
\bottomrule
\end{tabular}
\end{table*}

\begin{table*}[htbp]
\centering
\caption{Trade-off Between Coherence and Repetitiveness in BERTopic-generated Tweet Topics from TWEETS\_NYR.}
\label{tab:core-examples}
\begin{tabular}{l|p{8.1cm}|p{4.95cm}}
\toprule
\textbf{Topic ID} & \textbf{Examples (Topic Words)} & \textbf{Key Patterns} \\
\midrule
10 & \raggedright\textit{gainz, diet, fat, pretzel, \textbf{loseweight}, healthy, tweak, \textbf{dropthebabyweight}, \textbf{twinkie}, \textbf{shed}}
    & \makecell[l] {\raggedright Synonym pair:\\ \textit{\textbf{loseweight} $\leftrightarrow$ \textbf{shed}}\\
    \raggedright \textit{\textbf{dropthebabyweight} $\leftrightarrow$ \textbf{twinkie}}} \\
\midrule
11 & \raggedright\textit{\textbf{god}, \textbf{pray}, \textbf{jesus}, \textbf{prayer}, faith, bible, bless, miracle, church, hug}
    & \makecell[l] {\raggedright Synonym pairs:\\ \textit{\textbf{pray} $\leftrightarrow$ \textbf{prayer}}\\
    \raggedright \textit{\textbf{god} $\leftrightarrow$ \textbf{jesus}}} \\
\midrule
21 & \raggedright\textit{family, father, \textbf{mom}, bournemouth, resolutions, sheep, taxpayer, \textbf{mombiz}, sonoma, hangry} 
    & \makecell[l] {\raggedright Portmanteau:\\ \textit{\textbf{mombiz} $\leftrightarrow$ \textbf{mom}}} \\
\midrule
28 & \raggedright\textit{spend, debt, budget, \textbf{finance}, \textbf{account}, \textbf{wallet}, student, archive, \textbf{bank}, goodwill}
    & \makecell[l] {\raggedright Functional duplicates:\\ \textit{\textbf{account} $\leftrightarrow$ \textbf{wallet}}\\
    \raggedright Hierarchical link:\\ \textit{\textbf{bank} $\leftrightarrow$ \textbf{finance}}} \\
\midrule
34 & \raggedright\textit{\textbf{run}, mile, \textbf{marathon}, educator, iowa, runnerspace, racing, wyoming, marathontraine, marathontraining}
    & \makecell[l] {\raggedright Hierarchical link:\\ \textit{\textbf{marathon} $\leftrightarrow$ \textbf{run}}} \\
\midrule
39 & \raggedright\textit{\textbf{hashtag}, \textbf{hashtagoftheweek}, smoking, organize, hashtagofthe, hammer, medical, sew, recycle, broad}
    & \makecell[l] {\raggedright Hashtag redundancy:\\ \textit{\textbf{hashtagoftheweek} $\leftrightarrow$ \textbf{hashtag}}} \\
\bottomrule
\end{tabular}
\end{table*}

\subsubsection*{Outlier Detection Discrepancies}

Outlier detection is a crucial aspect of evaluating topic coherence, as it identifies semantically inconsistent words in a topic word set. Across the examples, outliers identified by the models often intersect but also reflect unique insights. 
Table \ref{table:example_outlier_detection} shows the examples of outlier detection discrepancies across different LLMs. 
Compared to the other two models, Mistral is more cautious in detecting outliers in topic words. On the contrary, Qwen is relatively more aggressive in detecting words with unclear semantic pointing from topic words and considering them as outliers.

\subsubsection*{Duplicate Concept Detection Contradictions}

The extracted duplicate pairs often differ significantly among the LLMs, showcasing varying thresholds for identifying conceptual overlap.
Table \ref{table:example_repetitiveness} shows that 
Mistral treats semantically related nouns (e.g., \textit{``christian''} and \textit{``church''}. In these collective nouns, there is intersection (e.g., \textit{``patient''} and \textit{``adult''}), and nouns that belong to the same category (e.g., \textit{``child''} and \textit{``adult''}) are conceptually identical. It has also had hallucinations (e.g., detecting a non-existent repetition of the word \textit{``customer''} for \textit{``client''} and a non-existent repetition of the word \textit{``email''} for \textit{``mail''}).
Llama treats grammatically related words (e.g., the verb ``search'' and its potential object ``package''), semantically opposite words (e.g., ``disease'' and ``health'') as conceptually identical.

\subsubsection*{BERTopic Coherence-repetitiveness Trade-off}
As shown in Table~\ref{tab:core-examples}.
BERTopic demonstrates strong thematic coherence across diverse topics (\textbf{ $C_\text{rate}{=3.0}$} in 35 out of 52 topics, e.g., Topic~11: \textit{god, pray, scripture}), effectively grouping related terms like \textit{marathon $\leftrightarrow$ run} (Topic~34).
However, repetitive patterns emerge through: (1) \textbf{synonym pairs} appearing in 43 out of 52 topics, particularly hierarchical relationships (\textit{bank $\leftrightarrow$ finance}) and functional equivalents (\textit{account $\leftrightarrow$ wallet} in Topic~28); (2) \textbf{informal language variants} like \texttt{gainz} (Topic~10) and \texttt{mombiz} (Topic~21) that fragment concepts. While maintaining semantic cohesion (87\% topics with $C_\text{rate}\geq2.0$), 32 out of 52 topics show elevated repetitiveness ($R_\text{rate}\geq2.0$), exacerbated by platform-specific conventions like hashtag variations (\texttt{hashtagoftheweek} $\leftrightarrow$ \texttt{hashtag} in Topic~39). This pattern highlights the model's effectiveness in capturing tweet semantics while struggling to consolidate informal lexical variations.

\section{Conclusion}
In this work, we introduced a purpose-oriented evaluation framework for topic modeling, leveraging LLMs to assess topic quality across multiple human-aligned dimensions. Our approach decomposes topic evaluation into interpretable subcomponents, including topic quality, including lexical validity, intra-topic semantic soundness, inter-topic structural soundness, and document-topic alignment soundness, and introduces dedicated LLM-based metrics for each. Through adversarial tests, correlation analysis, and large-scale empirical comparisons across datasets, topic models, and LLMs, we demonstrate that these metrics offer fine-grained diagnostic insights that complement and surpass traditional automated scores.  The findings highlight both the strengths and trade-offs of different topic modeling paradigms, illustrating how LLMs can serve as flexible evaluators capable of capturing nuanced aspects of topic quality. Our results also reveal systematic patterns in model performance across domains, as well as internal correlations among evaluation signals, laying the groundwork for more principled model selection and quality assurance in real-world text analysis pipelines.  By aligning evaluation with real-world purposes such as content understanding and document organization, this work contributes toward more meaningful and adaptive quality assessment in topic modeling. Future research may further explore the integration of human-in-the-loop feedback, extending the framework to multilingual corpora, and refining LLM-based prompts for domain-specific tasks.

{Addressing the evaluation of extremely long documents remains an open challenge. Future work could investigate the effectiveness of larger-context models or alternative approaches such as document segmentation, summarization, or key passage extraction.}

\bmhead{Acknowledgements}
Computing resources were provided by TIB – Leibniz Information Centre for Science and Technology.

\section*{Statements and Declarations}

\subsection*{Funding}
This work is jointly supported by the ``HybrInt - Hybrid Intelligence through Interpretable AI in Machine Perception and Interaction'' project (Zukunft Nds, Niedersächsisches Ministerium für Wissenschaft, Grant ID: ZN4219) and the SCINEXT project (BMBF, German Federal Ministry of Education and Research, Grant ID: 01lS22070)

\subsection*{Conflict of interest/Competing interests}
The authors declare that they have no known competing financial interests or personal relationships that could have appeared to influence the work reported in this paper.

\subsection*{Data Availability}
The datasets analyzed in this study are all publicly available. The preprocessed AGRIS dataset is provided in our public GitHub repository. The 20 Newsgroups and TWEETS\_NYR datasets were obtained from existing open sources, with their respective access links provided in footnotes within the manuscript. 

\subsection*{Code availability}
The codebase associated with this research is openly available on GitHub (\url{https://github.com/zhiyintan/topic-model-LLMjudgment}).


\bibliography{sn-bibliography}


\begin{thebibliography}{65}
\ifx \bisbn   \undefined \def \bisbn  #1{ISBN #1}\fi
\ifx \binits  \undefined \def \binits#1{#1}\fi
\ifx \bauthor  \undefined \def \bauthor#1{#1}\fi
\ifx \batitle  \undefined \def \batitle#1{#1}\fi
\ifx \bjtitle  \undefined \def \bjtitle#1{#1}\fi
\ifx \bvolume  \undefined \def \bvolume#1{\textbf{#1}}\fi
\ifx \byear  \undefined \def \byear#1{#1}\fi
\ifx \bissue  \undefined \def \bissue#1{#1}\fi
\ifx \bfpage  \undefined \def \bfpage#1{#1}\fi
\ifx \blpage  \undefined \def \blpage #1{#1}\fi
\ifx \burl  \undefined \def \burl#1{\textsf{#1}}\fi
\ifx \doiurl  \undefined \def \doiurl#1{\url{https://doi.org/#1}}\fi
\ifx \betal  \undefined \def \betal{\textit{et al.}}\fi
\ifx \binstitute  \undefined \def \binstitute#1{#1}\fi
\ifx \binstitutionaled  \undefined \def \binstitutionaled#1{#1}\fi
\ifx \bctitle  \undefined \def \bctitle#1{#1}\fi
\ifx \beditor  \undefined \def \beditor#1{#1}\fi
\ifx \bpublisher  \undefined \def \bpublisher#1{#1}\fi
\ifx \bbtitle  \undefined \def \bbtitle#1{#1}\fi
\ifx \bedition  \undefined \def \bedition#1{#1}\fi
\ifx \bseriesno  \undefined \def \bseriesno#1{#1}\fi
\ifx \blocation  \undefined \def \blocation#1{#1}\fi
\ifx \bsertitle  \undefined \def \bsertitle#1{#1}\fi
\ifx \bsnm \undefined \def \bsnm#1{#1}\fi
\ifx \bsuffix \undefined \def \bsuffix#1{#1}\fi
\ifx \bparticle \undefined \def \bparticle#1{#1}\fi
\ifx \barticle \undefined \def \barticle#1{#1}\fi
\bibcommenthead
\ifx \bconfdate \undefined \def \bconfdate #1{#1}\fi
\ifx \botherref \undefined \def \botherref #1{#1}\fi
\ifx \url \undefined \def \url#1{\textsf{#1}}\fi
\ifx \bchapter \undefined \def \bchapter#1{#1}\fi
\ifx \bbook \undefined \def \bbook#1{#1}\fi
\ifx \bcomment \undefined \def \bcomment#1{#1}\fi
\ifx \oauthor \undefined \def \oauthor#1{#1}\fi
\ifx \citeauthoryear \undefined \def \citeauthoryear#1{#1}\fi
\ifx \endbibitem  \undefined \def \endbibitem {}\fi
\ifx \bconflocation  \undefined \def \bconflocation#1{#1}\fi
\ifx \arxivurl  \undefined \def \arxivurl#1{\textsf{#1}}\fi
\csname PreBibitemsHook\endcsname

\bibitem[\protect\citeauthoryear{Blei et~al.}{2003}]{blei2003LDA}
\begin{barticle}
\bauthor{\bsnm{Blei}, \binits{D.M.}},
\bauthor{\bsnm{Ng}, \binits{A.Y.}},
\bauthor{\bsnm{Jordan}, \binits{M.I.}}:
\batitle{Latent dirichlet allocation}.
\bjtitle{Journal of machine Learning research}
\bvolume{3}(\bissue{Jan}),
\bfpage{993}--\blpage{1022}
(\byear{2003})
\end{barticle}
\endbibitem

\bibitem[\protect\citeauthoryear{Wang and Blei}{2011}]{wang2011CTR}
\begin{bchapter}
\bauthor{\bsnm{Wang}, \binits{C.}},
\bauthor{\bsnm{Blei}, \binits{D.M.}}:
\bctitle{Collaborative topic modeling for recommending scientific articles}.
In: \bbtitle{Proceedings of the 17th ACM SIGKDD International Conference on Knowledge Discovery and Data Mining}.
\bsertitle{KDD '11},
pp. \bfpage{448}--\blpage{456}.
\bpublisher{Association for Computing Machinery},
\blocation{New York, NY, USA}
(\byear{2011}).
\doiurl{10.1145/2020408.2020480} .
\burl{https://doi.org/10.1145/2020408.2020480}
\end{bchapter}
\endbibitem

\bibitem[\protect\citeauthoryear{Dwivedi et~al.}{2023}]{dwivedi-2023-evolutionAI}
\begin{barticle}
\bauthor{\bsnm{Dwivedi}, \binits{Y.K.}},
\bauthor{\bsnm{Sharma}, \binits{A.}},
\bauthor{\bsnm{Rana}, \binits{N.P.}},
\bauthor{\bsnm{Giannakis}, \binits{M.}},
\bauthor{\bsnm{Goel}, \binits{P.}},
\bauthor{\bsnm{Dutot}, \binits{V.}}:
\batitle{Evolution of artificial intelligence research in technological forecasting and social change: Research topics, trends, and future directions}.
\bjtitle{Technological Forecasting and Social Change}
\bvolume{192},
\bfpage{122579}
(\byear{2023})
\end{barticle}
\endbibitem

\bibitem[\protect\citeauthoryear{R\"{o}der et~al.}{2015}]{roder2015CV}
\begin{bchapter}
\bauthor{\bsnm{R\"{o}der}, \binits{M.}},
\bauthor{\bsnm{Both}, \binits{A.}},
\bauthor{\bsnm{Hinneburg}, \binits{A.}}:
\bctitle{Exploring the space of topic coherence measures}.
In: \bbtitle{Proceedings of the Eighth ACM International Conference on Web Search and Data Mining}.
\bsertitle{WSDM '15},
pp. \bfpage{399}--\blpage{408}.
\bpublisher{Association for Computing Machinery},
\blocation{New York, NY, USA}
(\byear{2015}).
\doiurl{10.1145/2684822.2685324} .
\burl{https://doi.org/10.1145/2684822.2685324}
\end{bchapter}
\endbibitem

\bibitem[\protect\citeauthoryear{Dieng et~al.}{2020}]{dieng2020ETM}
\begin{barticle}
\bauthor{\bsnm{Dieng}, \binits{A.B.}},
\bauthor{\bsnm{Ruiz}, \binits{F.J.R.}},
\bauthor{\bsnm{Blei}, \binits{D.M.}}:
\batitle{Topic modeling in embedding spaces}.
\bjtitle{Transactions of the Association for Computational Linguistics}
\bvolume{8},
\bfpage{439}--\blpage{453}
(\byear{2020})
\doiurl{10.1162/tacl_a_00325}
\end{barticle}
\endbibitem

\bibitem[\protect\citeauthoryear{Stammbach et~al.}{2023}]{stammbach2023revisitingLLM}
\begin{bchapter}
\bauthor{\bsnm{Stammbach}, \binits{D.}},
\bauthor{\bsnm{Zouhar}, \binits{V.}},
\bauthor{\bsnm{Hoyle}, \binits{A.}},
\bauthor{\bsnm{Sachan}, \binits{M.}},
\bauthor{\bsnm{Ash}, \binits{E.}}:
\bctitle{Revisiting automated topic model evaluation with large language models}.
In: \beditor{\bsnm{Bouamor}, \binits{H.}},
\beditor{\bsnm{Pino}, \binits{J.}},
\beditor{\bsnm{Bali}, \binits{K.}} (eds.)
\bbtitle{Proceedings of the 2023 Conference on Empirical Methods in Natural Language Processing},
pp. \bfpage{9348}--\blpage{9357}.
\bpublisher{Association for Computational Linguistics},
\blocation{Singapore}
(\byear{2023}).
\doiurl{10.18653/v1/2023.emnlp-main.581} .
\burl{https://aclanthology.org/2023.emnlp-main.581}
\end{bchapter}
\endbibitem

\bibitem[\protect\citeauthoryear{Tan and D'Souza}{2025}]{Tan2025Bridging}
\begin{bchapter}
\bauthor{\bsnm{Tan}, \binits{Z.}},
\bauthor{\bsnm{D'Souza}, \binits{J.}}:
\bctitle{{Bridging the Evaluation Gap: Leveraging Large Language Models for Topic Model Evaluation}}.
In: \beditor{\bsnm{Cornia}, \binits{M.}},
\beditor{\bsnm{Nunzio}, \binits{G.M.D.}},
\beditor{\bsnm{Firmani}, \binits{D.}},
\beditor{\bsnm{Mizzaro}, \binits{S.}},
\beditor{\bsnm{Serra}, \binits{G.}},
\beditor{\bsnm{Tonelli}, \binits{S.}},
\beditor{\bsnm{Tremamunno}, \binits{A.}} (eds.)
\bbtitle{Proceedings of the 21st Conference on Information and Research Science Connecting to Digital and Library Science (IRCDL 2025)}.
\bsertitle{CEUR Workshop Proceedings},
vol. \bseriesno{3937}.
\bpublisher{CEUR-WS.org},
\blocation{Udine, Italy}
(\byear{2025}).
\burl{http://ceur-ws.org/Vol-3937}
\end{bchapter}
\endbibitem

\bibitem[\protect\citeauthoryear{Papadimitriou et~al.}{1998}]{papadimitriou1998LSI}
\begin{bchapter}
\bauthor{\bsnm{Papadimitriou}, \binits{C.H.}},
\bauthor{\bsnm{Tamaki}, \binits{H.}},
\bauthor{\bsnm{Raghavan}, \binits{P.}},
\bauthor{\bsnm{Vempala}, \binits{S.}}:
\bctitle{Latent semantic indexing: A probabilistic analysis}.
In: \bbtitle{Proceedings of the Seventeenth ACM SIGACT-SIGMOD-SIGART Symposium on Principles of Database Systems},
pp. \bfpage{159}--\blpage{168}
(\byear{1998})
\end{bchapter}
\endbibitem

\bibitem[\protect\citeauthoryear{Hofmann}{1999}]{hofmann1999pLSI}
\begin{bchapter}
\bauthor{\bsnm{Hofmann}, \binits{T.}}:
\bctitle{Probabilistic latent semantic indexing}.
In: \bbtitle{Proceedings of the 22nd Annual International ACM SIGIR Conference on Research and Development in Information Retrieval}
(\byear{1999}).
\burl{https://sigir.org/wp-content/uploads/2017/06/p211.pdf}
\end{bchapter}
\endbibitem

\bibitem[\protect\citeauthoryear{Blei and Lafferty}{2007}]{blei2007CTM}
\begin{barticle}
\bauthor{\bsnm{Blei}, \binits{D.M.}},
\bauthor{\bsnm{Lafferty}, \binits{J.D.}}:
\batitle{A correlated topic model of science}.
\bjtitle{The Annals of Applied Statistics}
\bvolume{1},
\bfpage{17}--\blpage{35}
(\byear{2007})
\end{barticle}
\endbibitem

\bibitem[\protect\citeauthoryear{Newman et~al.}{2009}]{newman2009distributed}
\begin{botherref}
\oauthor{\bsnm{Newman}, \binits{D.}},
\oauthor{\bsnm{Asuncion}, \binits{A.}},
\oauthor{\bsnm{Smyth}, \binits{P.}},
\oauthor{\bsnm{Welling}, \binits{M.}}:
Distributed algorithms for topic models.
Journal of Machine Learning Research
\textbf{10}(8)
(2009)
\end{botherref}
\endbibitem

\bibitem[\protect\citeauthoryear{Chang et~al.}{2009}]{chang2009HumanTeaLeaves}
\begin{bchapter}
\bauthor{\bsnm{Chang}, \binits{J.}},
\bauthor{\bsnm{Boyd-Graber}, \binits{J.}},
\bauthor{\bsnm{Gerrish}, \binits{S.}},
\bauthor{\bsnm{Wang}, \binits{C.}},
\bauthor{\bsnm{Blei}, \binits{D.M.}}:
\bctitle{Reading tea leaves: how humans interpret topic models}.
In: \bbtitle{Proceedings of the 23rd International Conference on Neural Information Processing Systems}.
\bsertitle{NIPS'09},
pp. \bfpage{288}--\blpage{296}.
\bpublisher{Curran Associates Inc.},
\blocation{Red Hook, NY, USA}
(\byear{2009}).
\burl{https://dl.acm.org/doi/10.5555/2984093.2984126}
\end{bchapter}
\endbibitem

\bibitem[\protect\citeauthoryear{Mimno et~al.}{2011}]{mimno2011optimizing}
\begin{bchapter}
\bauthor{\bsnm{Mimno}, \binits{D.}},
\bauthor{\bsnm{Wallach}, \binits{H.}},
\bauthor{\bsnm{Talley}, \binits{E.}},
\bauthor{\bsnm{Leenders}, \binits{M.}},
\bauthor{\bsnm{McCallum}, \binits{A.}}:
\bctitle{Optimizing semantic coherence in topic models}.
In: \beditor{\bsnm{Barzilay}, \binits{R.}},
\beditor{\bsnm{Johnson}, \binits{M.}} (eds.)
\bbtitle{Proceedings of the 2011 Conference on Empirical Methods in Natural Language Processing},
pp. \bfpage{262}--\blpage{272}.
\bpublisher{Association for Computational Linguistics},
\blocation{Edinburgh, Scotland, UK.}
(\byear{2011}).
\burl{https://aclanthology.org/D11-1024}
\end{bchapter}
\endbibitem

\bibitem[\protect\citeauthoryear{Newman et~al.}{2010}]{Newman2010TMforDL}
\begin{bchapter}
\bauthor{\bsnm{Newman}, \binits{D.}},
\bauthor{\bsnm{Noh}, \binits{Y.}},
\bauthor{\bsnm{Talley}, \binits{E.}},
\bauthor{\bsnm{Karimi}, \binits{S.}},
\bauthor{\bsnm{Baldwin}, \binits{T.}}:
\bctitle{Evaluating topic models for digital libraries}.
In: \bbtitle{Proceedings of the 10th Annual Joint Conference on Digital Libraries}.
\bsertitle{JCDL '10},
pp. \bfpage{215}--\blpage{224}.
\bpublisher{Association for Computing Machinery},
\blocation{New York, NY, USA}
(\byear{2010}).
\doiurl{10.1145/1816123.1816156} .
\burl{https://doi.org/10.1145/1816123.1816156}
\end{bchapter}
\endbibitem

\bibitem[\protect\citeauthoryear{Bouma}{2009}]{bouma2009NPMI}
\begin{barticle}
\bauthor{\bsnm{Bouma}, \binits{G.}}:
\batitle{Normalized (pointwise) mutual information in collocation extraction}.
\bjtitle{Proceedings of GSCL}
\bvolume{30},
\bfpage{31}--\blpage{40}
(\byear{2009})
\end{barticle}
\endbibitem

\bibitem[\protect\citeauthoryear{Srivastava and Sutton}{2017}]{srivastava2017ProdLDA-AVITM}
\begin{bchapter}
\bauthor{\bsnm{Srivastava}, \binits{A.}},
\bauthor{\bsnm{Sutton}, \binits{C.}}:
\bctitle{Autoencoding variational inference for topic models}.
In: \bbtitle{International Conference on Learning Representations}
(\byear{2017}).
\burl{https://openreview.net/forum?id=BybtVK9lg}
\end{bchapter}
\endbibitem

\bibitem[\protect\citeauthoryear{Card et~al.}{2018}]{card2018SCHOLAR}
\begin{bchapter}
\bauthor{\bsnm{Card}, \binits{D.}},
\bauthor{\bsnm{Tan}, \binits{C.}},
\bauthor{\bsnm{Smith}, \binits{N.A.}}:
\bctitle{Neural models for documents with metadata}.
In: \beditor{\bsnm{Gurevych}, \binits{I.}},
\beditor{\bsnm{Miyao}, \binits{Y.}} (eds.)
\bbtitle{Proceedings of the 56th Annual Meeting of the Association for Computational Linguistics (Volume 1: Long Papers)},
pp. \bfpage{2031}--\blpage{2040}.
\bpublisher{Association for Computational Linguistics},
\blocation{Melbourne, Australia}
(\byear{2018}).
\doiurl{10.18653/v1/P18-1189} .
\burl{https://aclanthology.org/P18-1189}
\end{bchapter}
\endbibitem

\bibitem[\protect\citeauthoryear{Wang et~al.}{2019}]{wang2019ATM}
\begin{barticle}
\bauthor{\bsnm{Wang}, \binits{R.}},
\bauthor{\bsnm{Zhou}, \binits{D.}},
\bauthor{\bsnm{He}, \binits{Y.}}:
\batitle{Atm: Adversarial-neural topic model}.
\bjtitle{Information Processing \& Management}
\bvolume{56}(\bissue{6}),
\bfpage{102098}
(\byear{2019})
\doiurl{10.1016/j.ipm.2019.102098}
\end{barticle}
\endbibitem

\bibitem[\protect\citeauthoryear{Wang et~al.}{2020}]{wang2020BAT}
\begin{bchapter}
\bauthor{\bsnm{Wang}, \binits{R.}},
\bauthor{\bsnm{Hu}, \binits{X.}},
\bauthor{\bsnm{Zhou}, \binits{D.}},
\bauthor{\bsnm{He}, \binits{Y.}},
\bauthor{\bsnm{Xiong}, \binits{Y.}},
\bauthor{\bsnm{Ye}, \binits{C.}},
\bauthor{\bsnm{Xu}, \binits{H.}}:
\bctitle{Neural topic modeling with bidirectional adversarial training}.
In: \beditor{\bsnm{Jurafsky}, \binits{D.}},
\beditor{\bsnm{Chai}, \binits{J.}},
\beditor{\bsnm{Schluter}, \binits{N.}},
\beditor{\bsnm{Tetreault}, \binits{J.}} (eds.)
\bbtitle{Proceedings of the 58th Annual Meeting of the Association for Computational Linguistics},
pp. \bfpage{340}--\blpage{350}.
\bpublisher{Association for Computational Linguistics},
\blocation{Online}
(\byear{2020}).
\doiurl{10.18653/v1/2020.acl-main.32} .
\burl{https://aclanthology.org/2020.acl-main.32}
\end{bchapter}
\endbibitem

\bibitem[\protect\citeauthoryear{Bianchi et~al.}{2021}]{bianchi2021CombinedTM}
\begin{bchapter}
\bauthor{\bsnm{Bianchi}, \binits{F.}},
\bauthor{\bsnm{Terragni}, \binits{S.}},
\bauthor{\bsnm{Hovy}, \binits{D.}}:
\bctitle{Pre-training is a hot topic: Contextualized document embeddings improve topic coherence}.
In: \beditor{\bsnm{Zong}, \binits{C.}},
\beditor{\bsnm{Xia}, \binits{F.}},
\beditor{\bsnm{Li}, \binits{W.}},
\beditor{\bsnm{Navigli}, \binits{R.}} (eds.)
\bbtitle{Proceedings of the 59th Annual Meeting of the Association for Computational Linguistics and the 11th International Joint Conference on Natural Language Processing (Volume 2: Short Papers)},
pp. \bfpage{759}--\blpage{766}.
\bpublisher{Association for Computational Linguistics},
\blocation{Online}
(\byear{2021}).
\doiurl{10.18653/v1/2021.acl-short.96} .
\burl{https://aclanthology.org/2021.acl-short.96}
\end{bchapter}
\endbibitem

\bibitem[\protect\citeauthoryear{Wu et~al.}{2023}]{wu2023ECRTM}
\begin{bchapter}
\bauthor{\bsnm{Wu}, \binits{X.}},
\bauthor{\bsnm{Dong}, \binits{X.}},
\bauthor{\bsnm{Nguyen}, \binits{T.T.}},
\bauthor{\bsnm{Luu}, \binits{A.T.}}:
\bctitle{Effective neural topic modeling with embedding clustering regularization}.
In: \bbtitle{International Conference on Machine Learning},
pp. \bfpage{37335}--\blpage{37357}
(\byear{2023}).
\bcomment{PMLR}
\end{bchapter}
\endbibitem

\bibitem[\protect\citeauthoryear{Schnabel et~al.}{2015}]{schnabel2015EmbeddingEvalTM}
\begin{bchapter}
\bauthor{\bsnm{Schnabel}, \binits{T.}},
\bauthor{\bsnm{Labutov}, \binits{I.}},
\bauthor{\bsnm{Mimno}, \binits{D.}},
\bauthor{\bsnm{Joachims}, \binits{T.}}:
\bctitle{Evaluation methods for unsupervised word embeddings}.
In: \beditor{\bsnm{M{\`a}rquez}, \binits{L.}},
\beditor{\bsnm{Callison-Burch}, \binits{C.}},
\beditor{\bsnm{Su}, \binits{J.}} (eds.)
\bbtitle{Proceedings of the 2015 Conference on Empirical Methods in Natural Language Processing},
pp. \bfpage{298}--\blpage{307}.
\bpublisher{Association for Computational Linguistics},
\blocation{Lisbon, Portugal}
(\byear{2015}).
\doiurl{10.18653/v1/D15-1036} .
\burl{https://aclanthology.org/D15-1036}
\end{bchapter}
\endbibitem

\bibitem[\protect\citeauthoryear{Nikolenko}{2016}]{Nikolenko2016EmbeddingEvalTM}
\begin{bchapter}
\bauthor{\bsnm{Nikolenko}, \binits{S.I.}}:
\bctitle{Topic quality metrics based on distributed word representations}.
In: \bbtitle{Proceedings of the 39th International ACM SIGIR Conference on Research and Development in Information Retrieval}.
\bsertitle{SIGIR '16},
pp. \bfpage{1029}--\blpage{1032}.
\bpublisher{Association for Computing Machinery},
\blocation{New York, NY, USA}
(\byear{2016}).
\doiurl{10.1145/2911451.2914720} .
\burl{https://doi.org/10.1145/2911451.2914720}
\end{bchapter}
\endbibitem

\bibitem[\protect\citeauthoryear{Ding et~al.}{2018}]{ding2018coherenceNTM}
\begin{bchapter}
\bauthor{\bsnm{Ding}, \binits{R.}},
\bauthor{\bsnm{Nallapati}, \binits{R.}},
\bauthor{\bsnm{Xiang}, \binits{B.}}:
\bctitle{Coherence-aware neural topic modeling}.
In: \beditor{\bsnm{Riloff}, \binits{E.}},
\beditor{\bsnm{Chiang}, \binits{D.}},
\beditor{\bsnm{Hockenmaier}, \binits{J.}},
\beditor{\bsnm{Tsujii}, \binits{J.}} (eds.)
\bbtitle{Proceedings of the 2018 Conference on Empirical Methods in Natural Language Processing},
pp. \bfpage{830}--\blpage{836}.
\bpublisher{Association for Computational Linguistics},
\blocation{Brussels, Belgium}
(\byear{2018}).
\doiurl{10.18653/v1/D18-1096} .
\burl{https://aclanthology.org/D18-1096}
\end{bchapter}
\endbibitem

\bibitem[\protect\citeauthoryear{Sia et~al.}{2020}]{sia2020tired-cluster}
\begin{bchapter}
\bauthor{\bsnm{Sia}, \binits{S.}},
\bauthor{\bsnm{Dalmia}, \binits{A.}},
\bauthor{\bsnm{Mielke}, \binits{S.J.}}:
\bctitle{Tired of topic models? clusters of pretrained word embeddings make for fast and good topics too!}
In: \beditor{\bsnm{Webber}, \binits{B.}},
\beditor{\bsnm{Cohn}, \binits{T.}},
\beditor{\bsnm{He}, \binits{Y.}},
\beditor{\bsnm{Liu}, \binits{Y.}} (eds.)
\bbtitle{Proceedings of the 2020 Conference on Empirical Methods in Natural Language Processing (EMNLP)},
pp. \bfpage{1728}--\blpage{1736}.
\bpublisher{Association for Computational Linguistics},
\blocation{Online}
(\byear{2020}).
\doiurl{10.18653/v1/2020.emnlp-main.135} .
\burl{https://aclanthology.org/2020.emnlp-main.135}
\end{bchapter}
\endbibitem

\bibitem[\protect\citeauthoryear{Grootendorst}{2022}]{grootendorst2022bertopic}
\begin{botherref}
\oauthor{\bsnm{Grootendorst}, \binits{M.}}:
Bertopic: Neural topic modeling with a class-based tf-idf procedure.
arXiv preprint arXiv:2203.05794
(2022)
\end{botherref}
\endbibitem

\bibitem[\protect\citeauthoryear{Burkhardt and Kramer}{2019}]{Burkhardt2019TopicRedundancy}
\begin{barticle}
\bauthor{\bsnm{Burkhardt}, \binits{S.}},
\bauthor{\bsnm{Kramer}, \binits{S.}}:
\batitle{Decoupling sparsity and smoothness in the dirichlet variational autoencoder topic model}.
\bjtitle{Journal of Machine Learning Research}
\bvolume{20}(\bissue{131}),
\bfpage{1}--\blpage{27}
(\byear{2019})
\end{barticle}
\endbibitem

\bibitem[\protect\citeauthoryear{Nan et~al.}{2019}]{nan2019TopicUniqueness}
\begin{bchapter}
\bauthor{\bsnm{Nan}, \binits{F.}},
\bauthor{\bsnm{Ding}, \binits{R.}},
\bauthor{\bsnm{Nallapati}, \binits{R.}},
\bauthor{\bsnm{Xiang}, \binits{B.}}:
\bctitle{Topic modeling with {W}asserstein autoencoders}.
In: \beditor{\bsnm{Korhonen}, \binits{A.}},
\beditor{\bsnm{Traum}, \binits{D.}},
\beditor{\bsnm{M{\`a}rquez}, \binits{L.}} (eds.)
\bbtitle{Proceedings of the 57th Annual Meeting of the Association for Computational Linguistics},
pp. \bfpage{6345}--\blpage{6381}.
\bpublisher{Association for Computational Linguistics},
\blocation{Florence, Italy}
(\byear{2019}).
\doiurl{10.18653/v1/P19-1640} .
\burl{https://aclanthology.org/P19-1640}
\end{bchapter}
\endbibitem

\bibitem[\protect\citeauthoryear{Bianchi et~al.}{2021}]{bianchi2021EmbedingCentroidTD}
\begin{bchapter}
\bauthor{\bsnm{Bianchi}, \binits{F.}},
\bauthor{\bsnm{Terragni}, \binits{S.}},
\bauthor{\bsnm{Hovy}, \binits{D.}},
\bauthor{\bsnm{Nozza}, \binits{D.}},
\bauthor{\bsnm{Fersini}, \binits{E.}}:
\bctitle{Cross-lingual contextualized topic models with zero-shot learning}.
In: \beditor{\bsnm{Merlo}, \binits{P.}},
\beditor{\bsnm{Tiedemann}, \binits{J.}},
\beditor{\bsnm{Tsarfaty}, \binits{R.}} (eds.)
\bbtitle{Proceedings of the 16th Conference of the European Chapter of the Association for Computational Linguistics: Main Volume},
pp. \bfpage{1676}--\blpage{1683}.
\bpublisher{Association for Computational Linguistics},
\blocation{Online}
(\byear{2021}).
\doiurl{10.18653/v1/2021.eacl-main.143} .
\burl{https://aclanthology.org/2021.eacl-main.143}
\end{bchapter}
\endbibitem

\bibitem[\protect\citeauthoryear{Terragni et~al.}{2021}]{Terragni2021EmbedingTD}
\begin{bchapter}
\bauthor{\bsnm{Terragni}, \binits{S.}},
\bauthor{\bsnm{Fersini}, \binits{E.}},
\bauthor{\bsnm{Messina}, \binits{E.}}:
\bctitle{Word embedding-based topic similarity measures}.
In: \beditor{\bsnm{M{\'e}tais}, \binits{E.}},
\beditor{\bsnm{Meziane}, \binits{F.}},
\beditor{\bsnm{Horacek}, \binits{H.}},
\beditor{\bsnm{Kapetanios}, \binits{E.}} (eds.)
\bbtitle{Natural Language Processing and Information Systems},
pp. \bfpage{33}--\blpage{45}.
\bpublisher{Springer},
\blocation{Cham}
(\byear{2021})
\end{bchapter}
\endbibitem

\bibitem[\protect\citeauthoryear{Bhatia et~al.}{2017}]{bhatia2017DocumentTopicEval}
\begin{bchapter}
\bauthor{\bsnm{Bhatia}, \binits{S.}},
\bauthor{\bsnm{Lau}, \binits{J.H.}},
\bauthor{\bsnm{Baldwin}, \binits{T.}}:
\bctitle{An automatic approach for document-level topic model evaluation}.
In: \beditor{\bsnm{Levy}, \binits{R.}},
\beditor{\bsnm{Specia}, \binits{L.}} (eds.)
\bbtitle{Proceedings of the 21st Conference on Computational Natural Language Learning ({C}o{NLL} 2017)},
pp. \bfpage{206}--\blpage{215}.
\bpublisher{Association for Computational Linguistics},
\blocation{Vancouver, Canada}
(\byear{2017}).
\doiurl{10.18653/v1/K17-1022} .
\burl{https://aclanthology.org/K17-1022}
\end{bchapter}
\endbibitem

\bibitem[\protect\citeauthoryear{Bhatia et~al.}{2018}]{bhatia2018topicIntrusionEval}
\begin{bchapter}
\bauthor{\bsnm{Bhatia}, \binits{S.}},
\bauthor{\bsnm{Lau}, \binits{J.H.}},
\bauthor{\bsnm{Baldwin}, \binits{T.}}:
\bctitle{Topic intrusion for automatic topic model evaluation}.
In: \beditor{\bsnm{Riloff}, \binits{E.}},
\beditor{\bsnm{Chiang}, \binits{D.}},
\beditor{\bsnm{Hockenmaier}, \binits{J.}},
\beditor{\bsnm{Tsujii}, \binits{J.}} (eds.)
\bbtitle{Proceedings of the 2018 Conference on Empirical Methods in Natural Language Processing},
pp. \bfpage{844}--\blpage{849}.
\bpublisher{Association for Computational Linguistics},
\blocation{Brussels, Belgium}
(\byear{2018}).
\doiurl{10.18653/v1/D18-1098} .
\burl{https://aclanthology.org/D18-1098}
\end{bchapter}
\endbibitem

\bibitem[\protect\citeauthoryear{Korenčić et~al.}{2018}]{Korencic2018DocumenCoherence}
\begin{barticle}
\bauthor{\bsnm{Korenčić}, \binits{D.}},
\bauthor{\bsnm{Ristov}, \binits{S.}},
\bauthor{\bsnm{Šnajder}, \binits{J.}}:
\batitle{Document-based topic coherence measures for news media text}.
\bjtitle{Expert Systems with Applications}
\bvolume{114},
\bfpage{357}--\blpage{373}
(\byear{2018})
\doiurl{10.1016/j.eswa.2018.07.063}
\end{barticle}
\endbibitem

\bibitem[\protect\citeauthoryear{Koren{\v{c}}i{\'c} et~al.}{2021}]{korenvcic2021CoverageEval}
\begin{barticle}
\bauthor{\bsnm{Koren{\v{c}}i{\'c}}, \binits{D.}},
\bauthor{\bsnm{Ristov}, \binits{S.}},
\bauthor{\bsnm{Repar}, \binits{J.}},
\bauthor{\bsnm{{\v{S}}najder}, \binits{J.}}:
\batitle{A topic coverage approach to evaluation of topic models}.
\bjtitle{IEEE access}
\bvolume{9},
\bfpage{123280}--\blpage{123312}
(\byear{2021})
\end{barticle}
\endbibitem

\bibitem[\protect\citeauthoryear{Rahimi et~al.}{2024}]{rahimi2024contextualizedCoherence}
\begin{bchapter}
\bauthor{\bsnm{Rahimi}, \binits{H.}},
\bauthor{\bsnm{Mimno}, \binits{D.}},
\bauthor{\bsnm{Hoover}, \binits{J.}},
\bauthor{\bsnm{Naacke}, \binits{H.}},
\bauthor{\bsnm{Constantin}, \binits{C.}},
\bauthor{\bsnm{Amann}, \binits{B.}}:
\bctitle{Contextualized topic coherence metrics}.
In: \beditor{\bsnm{Graham}, \binits{Y.}},
\beditor{\bsnm{Purver}, \binits{M.}} (eds.)
\bbtitle{Findings of the Association for Computational Linguistics: EACL 2024},
pp. \bfpage{1760}--\blpage{1773}.
\bpublisher{Association for Computational Linguistics},
\blocation{St. Julian{'}s, Malta}
(\byear{2024}).
\burl{https://aclanthology.org/2024.findings-eacl.123}
\end{bchapter}
\endbibitem

\bibitem[\protect\citeauthoryear{Yang et~al.}{2024}]{yang2024llmReadingTeaLeaves}
\begin{botherref}
\oauthor{\bsnm{Yang}, \binits{X.}},
\oauthor{\bsnm{Zhao}, \binits{H.}},
\oauthor{\bsnm{Phung}, \binits{D.}},
\oauthor{\bsnm{Buntine}, \binits{W.}},
\oauthor{\bsnm{Du}, \binits{L.}}:
Llm reading tea leaves: Automatically evaluating topic models with large language models.
arXiv preprint arXiv:2406.09008
(2024)
\end{botherref}
\endbibitem

\bibitem[\protect\citeauthoryear{Blei}{2012}]{blei2012PLDA}
\begin{barticle}
\bauthor{\bsnm{Blei}, \binits{D.M.}}:
\batitle{Probabilistic topic models}.
\bjtitle{Commun. ACM}
\bvolume{55}(\bissue{4}),
\bfpage{77}--\blpage{84}
(\byear{2012})
\doiurl{10.1145/2133806.2133826}
\end{barticle}
\endbibitem

\bibitem[\protect\citeauthoryear{Doogan and Buntine}{2021}]{doogan-buntine-2021-TopicTwaddle}
\begin{bchapter}
\bauthor{\bsnm{Doogan}, \binits{C.}},
\bauthor{\bsnm{Buntine}, \binits{W.}}:
\bctitle{Topic model or topic twaddle? re-evaluating semantic interpretability measures}.
In: \beditor{\bsnm{Toutanova}, \binits{K.}},
\beditor{\bsnm{Rumshisky}, \binits{A.}},
\beditor{\bsnm{Zettlemoyer}, \binits{L.}},
\beditor{\bsnm{Hakkani-Tur}, \binits{D.}},
\beditor{\bsnm{Beltagy}, \binits{I.}},
\beditor{\bsnm{Bethard}, \binits{S.}},
\beditor{\bsnm{Cotterell}, \binits{R.}},
\beditor{\bsnm{Chakraborty}, \binits{T.}},
\beditor{\bsnm{Zhou}, \binits{Y.}} (eds.)
\bbtitle{Proceedings of the 2021 Conference of the North American Chapter of the Association for Computational Linguistics: Human Language Technologies},
pp. \bfpage{3824}--\blpage{3848}.
\bpublisher{Association for Computational Linguistics},
\blocation{Online}
(\byear{2021}).
\doiurl{10.18653/v1/2021.naacl-main.300} .
\burl{https://aclanthology.org/2021.naacl-main.300/}
\end{bchapter}
\endbibitem

\bibitem[\protect\citeauthoryear{Griffiths and Steyvers}{2004}]{griffiths-2004-gibbs}
\begin{barticle}
\bauthor{\bsnm{Griffiths}, \binits{T.L.}},
\bauthor{\bsnm{Steyvers}, \binits{M.}}:
\batitle{Finding scientific topics}.
\bjtitle{Proceedings of the National Academy of Sciences}
\bvolume{101}(\bissue{suppl\_1}),
\bfpage{5228}--\blpage{5235}
(\byear{2004})
\doiurl{10.1073/pnas.0307752101}
\end{barticle}
\endbibitem

\bibitem[\protect\citeauthoryear{Lee et~al.}{2021}]{lee-2021-sustainability}
\begin{barticle}
\bauthor{\bsnm{Lee}, \binits{J.H.}},
\bauthor{\bsnm{Wood}, \binits{J.}},
\bauthor{\bsnm{Kim}, \binits{J.}}:
\batitle{Tracing the trends in sustainability and social media research using topic modeling}.
\bjtitle{Sustainability}
\bvolume{13}(\bissue{3}),
\bfpage{1269}
(\byear{2021})
\end{barticle}
\endbibitem

\bibitem[\protect\citeauthoryear{Mejia et~al.}{2021}]{mejia-2021-bibliometric}
\begin{barticle}
\bauthor{\bsnm{Mejia}, \binits{C.}},
\bauthor{\bsnm{Wu}, \binits{M.}},
\bauthor{\bsnm{Zhang}, \binits{Y.}},
\bauthor{\bsnm{Kajikawa}, \binits{Y.}}:
\batitle{Exploring topics in bibliometric research through citation networks and semantic analysis}.
\bjtitle{Frontiers in Research Metrics and Analytics}
\bvolume{6},
\bfpage{742311}
(\byear{2021})
\end{barticle}
\endbibitem

\bibitem[\protect\citeauthoryear{Koltsova and Koltcov}{2013}]{koltsova-2013-publicRussian}
\begin{barticle}
\bauthor{\bsnm{Koltsova}, \binits{O.}},
\bauthor{\bsnm{Koltcov}, \binits{S.}}:
\batitle{Mapping the public agenda with topic modeling: The case of the russian livejournal}.
\bjtitle{Policy \& Internet}
\bvolume{5}(\bissue{2}),
\bfpage{207}--\blpage{227}
(\byear{2013})
\end{barticle}
\endbibitem

\bibitem[\protect\citeauthoryear{de~Melo et~al.}{2021}]{deMelo-2021-covid19Brazil}
\begin{barticle}
\bauthor{\bsnm{Melo}, \binits{T.}},
\bauthor{\bsnm{Figueiredo}, \binits{C.M.}}, \betal:
\batitle{Comparing news articles and tweets about covid-19 in brazil: sentiment analysis and topic modeling approach}.
\bjtitle{JMIR Public Health and Surveillance}
\bvolume{7}(\bissue{2}),
\bfpage{24585}
(\byear{2021})
\end{barticle}
\endbibitem

\bibitem[\protect\citeauthoryear{Aletras and Stevenson}{2013}]{aletras2013TC}
\begin{bchapter}
\bauthor{\bsnm{Aletras}, \binits{N.}},
\bauthor{\bsnm{Stevenson}, \binits{M.}}:
\bctitle{Evaluating topic coherence using distributional semantics}.
In: \beditor{\bsnm{Koller}, \binits{A.}},
\beditor{\bsnm{Erk}, \binits{K.}} (eds.)
\bbtitle{Proceedings of the 10th International Conference on Computational Semantics ({IWCS} 2013) {--} Long Papers},
pp. \bfpage{13}--\blpage{22}.
\bpublisher{Association for Computational Linguistics},
\blocation{Potsdam, Germany}
(\byear{2013}).
\burl{https://aclanthology.org/W13-0102}
\end{bchapter}
\endbibitem

\bibitem[\protect\citeauthoryear{Newman et~al.}{2010}]{newman2010automaticHuman}
\begin{bchapter}
\bauthor{\bsnm{Newman}, \binits{D.}},
\bauthor{\bsnm{Lau}, \binits{J.H.}},
\bauthor{\bsnm{Grieser}, \binits{K.}},
\bauthor{\bsnm{Baldwin}, \binits{T.}}:
\bctitle{Automatic evaluation of topic coherence}.
In: \beditor{\bsnm{Kaplan}, \binits{R.}},
\beditor{\bsnm{Burstein}, \binits{J.}},
\beditor{\bsnm{Harper}, \binits{M.}},
\beditor{\bsnm{Penn}, \binits{G.}} (eds.)
\bbtitle{Human Language Technologies: The 2010 Annual Conference of the North {A}merican Chapter of the Association for Computational Linguistics},
pp. \bfpage{100}--\blpage{108}.
\bpublisher{Association for Computational Linguistics},
\blocation{Los Angeles, California}
(\byear{2010}).
\burl{https://aclanthology.org/N10-1012}
\end{bchapter}
\endbibitem

\bibitem[\protect\citeauthoryear{Lu et~al.}{2011}]{lu-2011-legalCluster}
\begin{bchapter}
\bauthor{\bsnm{Lu}, \binits{Q.}},
\bauthor{\bsnm{Conrad}, \binits{J.G.}},
\bauthor{\bsnm{Al-Kofahi}, \binits{K.}},
\bauthor{\bsnm{Keenan}, \binits{W.}}:
\bctitle{Legal document clustering with built-in topic segmentation}.
In: \bbtitle{Proceedings of the 20th ACM International Conference on Information and Knowledge Management},
pp. \bfpage{383}--\blpage{392}
(\byear{2011})
\end{bchapter}
\endbibitem

\bibitem[\protect\citeauthoryear{Aguiar et~al.}{2022}]{aguiar-2022-brazilianLawsuits}
\begin{bchapter}
\bauthor{\bsnm{Aguiar}, \binits{A.}},
\bauthor{\bsnm{Silveira}, \binits{R.}},
\bauthor{\bsnm{Furtado}, \binits{V.}},
\bauthor{\bsnm{Pinheiro}, \binits{V.}},
\bauthor{\bsnm{Neto}, \binits{J.A.M.}}:
\bctitle{Using topic modeling in classification of brazilian lawsuits}.
In: \bbtitle{International Conference on Computational Processing of the Portuguese Language},
pp. \bfpage{233}--\blpage{242}
(\byear{2022}).
\bcomment{Springer}
\end{bchapter}
\endbibitem

\bibitem[\protect\citeauthoryear{Tuarob et~al.}{2015}]{tuarob-2015-annotationEnvironment}
\begin{barticle}
\bauthor{\bsnm{Tuarob}, \binits{S.}},
\bauthor{\bsnm{Pouchard}, \binits{L.C.}},
\bauthor{\bsnm{Mitra}, \binits{P.}},
\bauthor{\bsnm{Giles}, \binits{C.L.}}:
\batitle{A generalized topic modeling approach for automatic document annotation}.
\bjtitle{International Journal on Digital Libraries}
\bvolume{16},
\bfpage{111}--\blpage{128}
(\byear{2015})
\end{barticle}
\endbibitem

\bibitem[\protect\citeauthoryear{Newman et~al.}{2007}]{newman-2007-metadataEnrichment}
\begin{bchapter}
\bauthor{\bsnm{Newman}, \binits{D.}},
\bauthor{\bsnm{Hagedorn}, \binits{K.}},
\bauthor{\bsnm{Chemudugunta}, \binits{C.}},
\bauthor{\bsnm{Smyth}, \binits{P.}}:
\bctitle{Subject metadata enrichment using statistical topic models}.
In: \bbtitle{Proceedings of the 7th ACM/IEEE-CS Joint Conference on Digital Libraries},
pp. \bfpage{366}--\blpage{375}
(\byear{2007})
\end{bchapter}
\endbibitem

\bibitem[\protect\citeauthoryear{Cain}{2016}]{cain-2016-accesslibrary}
\begin{barticle}
\bauthor{\bsnm{Cain}, \binits{J.O.}}:
\batitle{Using topic modeling to enhance access to library digital collections}.
\bjtitle{Journal of Web Librarianship}
\bvolume{10}(\bissue{3}),
\bfpage{210}--\blpage{225}
(\byear{2016})
\end{barticle}
\endbibitem

\bibitem[\protect\citeauthoryear{Ramage et~al.}{2011}]{ramage-2011-partiallyLabel}
\begin{bchapter}
\bauthor{\bsnm{Ramage}, \binits{D.}},
\bauthor{\bsnm{Manning}, \binits{C.D.}},
\bauthor{\bsnm{Dumais}, \binits{S.}}:
\bctitle{Partially labeled topic models for interpretable text mining}.
In: \bbtitle{Proceedings of the 17th ACM SIGKDD International Conference on Knowledge Discovery and Data Mining},
pp. \bfpage{457}--\blpage{465}
(\byear{2011})
\end{bchapter}
\endbibitem

\bibitem[\protect\citeauthoryear{Hingmire et~al.}{2013}]{hingmire-2013-documentClassification}
\begin{bchapter}
\bauthor{\bsnm{Hingmire}, \binits{S.}},
\bauthor{\bsnm{Chougule}, \binits{S.}},
\bauthor{\bsnm{Palshikar}, \binits{G.K.}},
\bauthor{\bsnm{Chakraborti}, \binits{S.}}:
\bctitle{Document classification by topic labeling}.
In: \bbtitle{Proceedings of the 36th International ACM SIGIR Conference on Research and Development in Information Retrieval},
pp. \bfpage{877}--\blpage{880}
(\byear{2013})
\end{bchapter}
\endbibitem

\bibitem[\protect\citeauthoryear{Nguyen et~al.}{2015}]{nguyen-2015-latentFeature}
\begin{barticle}
\bauthor{\bsnm{Nguyen}, \binits{D.Q.}},
\bauthor{\bsnm{Billingsley}, \binits{R.}},
\bauthor{\bsnm{Du}, \binits{L.}},
\bauthor{\bsnm{Johnson}, \binits{M.}}:
\batitle{Improving topic models with latent feature word representations}.
\bjtitle{Transactions of the Association for Computational Linguistics}
\bvolume{3},
\bfpage{299}--\blpage{313}
(\byear{2015})
\doiurl{10.1162/tacl_a_00140}
\end{barticle}
\endbibitem

\bibitem[\protect\citeauthoryear{Sch{\"u}tze et~al.}{2008}]{schutze-2008-NMI}
\begin{bbook}
\bauthor{\bsnm{Sch{\"u}tze}, \binits{H.}},
\bauthor{\bsnm{Manning}, \binits{C.D.}},
\bauthor{\bsnm{Raghavan}, \binits{P.}}:
\bbtitle{Introduction to Information Retrieval}
vol. \bseriesno{39}.
\bpublisher{Cambridge University Press Cambridge}, \blocation{???}
(\byear{2008})
\end{bbook}
\endbibitem

\bibitem[\protect\citeauthoryear{Zhao et~al.}{2021}]{zhao-2021-ntmOptimal}
\begin{bchapter}
\bauthor{\bsnm{Zhao}, \binits{H.}},
\bauthor{\bsnm{Phung}, \binits{D.}},
\bauthor{\bsnm{Huynh}, \binits{V.}},
\bauthor{\bsnm{Le}, \binits{T.}},
\bauthor{\bsnm{Buntine}, \binits{W.}}:
\bctitle{Neural topic model via optimal transport}.
In: \bbtitle{International Conference on Learning Representations}
(\byear{2021}).
\burl{https://openreview.net/forum?id=Oos98K9Lv-k}
\end{bchapter}
\endbibitem

\bibitem[\protect\citeauthoryear{Zhao et~al.}{2016}]{zhao-2016-hashtagRecommendation}
\begin{barticle}
\bauthor{\bsnm{Zhao}, \binits{F.}},
\bauthor{\bsnm{Zhu}, \binits{Y.}},
\bauthor{\bsnm{Jin}, \binits{H.}},
\bauthor{\bsnm{Yang}, \binits{L.T.}}:
\batitle{A personalized hashtag recommendation approach using lda-based topic model in microblog environment}.
\bjtitle{Future Generation Computer Systems}
\bvolume{65},
\bfpage{196}--\blpage{206}
(\byear{2016})
\end{barticle}
\endbibitem

\bibitem[\protect\citeauthoryear{Srivastava et~al.}{2022}]{srivastava-2022-singleSummarization}
\begin{barticle}
\bauthor{\bsnm{Srivastava}, \binits{R.}},
\bauthor{\bsnm{Singh}, \binits{P.}},
\bauthor{\bsnm{Rana}, \binits{K.}},
\bauthor{\bsnm{Kumar}, \binits{V.}}:
\batitle{A topic modeled unsupervised approach to single document extractive text summarization}.
\bjtitle{Knowledge-Based Systems}
\bvolume{246},
\bfpage{108636}
(\byear{2022})
\end{barticle}
\endbibitem

\bibitem[\protect\citeauthoryear{Nagwani}{2015}]{nagwani-2015-mapReduceSummarizing}
\begin{barticle}
\bauthor{\bsnm{Nagwani}, \binits{N.K.}}:
\batitle{Summarizing large text collection using topic modeling and clustering based on mapreduce framework}.
\bjtitle{Journal of Big Data}
\bvolume{2}(\bissue{1}),
\bfpage{6}
(\byear{2015})
\end{barticle}
\endbibitem

\bibitem[\protect\citeauthoryear{Salakhutdinov and Hinton}{2009}]{Hinton2009UndirectedTM}
\begin{bchapter}
\bauthor{\bsnm{Salakhutdinov}, \binits{R.}},
\bauthor{\bsnm{Hinton}, \binits{G.}}:
\bctitle{Replicated softmax: an undirected topic model}.
In: \bbtitle{Proceedings of the 23rd International Conference on Neural Information Processing Systems}.
\bsertitle{NIPS'09},
pp. \bfpage{1607}--\blpage{1614}.
\bpublisher{Curran Associates Inc.},
\blocation{Red Hook, NY, USA}
(\byear{2009})
\end{bchapter}
\endbibitem

\bibitem[\protect\citeauthoryear{Miao et~al.}{2017}]{miao2017GSM-GSB-RSB}
\begin{bchapter}
\bauthor{\bsnm{Miao}, \binits{Y.}},
\bauthor{\bsnm{Grefenstette}, \binits{E.}},
\bauthor{\bsnm{Blunsom}, \binits{P.}}:
\bctitle{Discovering discrete latent topics with neural variational inference}.
In: \beditor{\bsnm{Precup}, \binits{D.}},
\beditor{\bsnm{Teh}, \binits{Y.W.}} (eds.)
\bbtitle{Proceedings of the 34th International Conference on Machine Learning}.
\bsertitle{Proceedings of Machine Learning Research},
vol. \bseriesno{70},
pp. \bfpage{2410}--\blpage{2419}.
\bpublisher{PMLR}, \blocation{???}
(\byear{2017}).
\burl{http://proceedings.mlr.press/v70/miao17a.html}
\end{bchapter}
\endbibitem

\bibitem[\protect\citeauthoryear{Zhang et~al.}{2018}]{Zhang2018WHAI}
\begin{botherref}
\oauthor{\bsnm{Zhang}, \binits{H.}},
\oauthor{\bsnm{Chen}, \binits{B.}},
\oauthor{\bsnm{Guo}, \binits{D.}},
\oauthor{\bsnm{Zhou}, \binits{M.}}:
Whai: Weibull hybrid autoencoding inference for deep topic modeling.
arXiv preprint arXiv:1803.01328
(2018)
\end{botherref}
\endbibitem

\bibitem[\protect\citeauthoryear{Frohmann et~al.}{2024}]{frohmann-etal-2024-segment}
\begin{bchapter}
\bauthor{\bsnm{Frohmann}, \binits{M.}},
\bauthor{\bsnm{Sterner}, \binits{I.}},
\bauthor{\bsnm{Vuli{\'c}}, \binits{I.}},
\bauthor{\bsnm{Minixhofer}, \binits{B.}},
\bauthor{\bsnm{Schedl}, \binits{M.}}:
\bctitle{Segment any text: A universal approach for robust, efficient and adaptable sentence segmentation}.
In: \beditor{\bsnm{Al-Onaizan}, \binits{Y.}},
\beditor{\bsnm{Bansal}, \binits{M.}},
\beditor{\bsnm{Chen}, \binits{Y.-N.}} (eds.)
\bbtitle{Proceedings of the 2024 Conference on Empirical Methods in Natural Language Processing},
pp. \bfpage{11908}--\blpage{11941}.
\bpublisher{Association for Computational Linguistics},
\blocation{Miami, Florida, USA}
(\byear{2024}).
\burl{https://aclanthology.org/2024.emnlp-main.665}
\end{bchapter}
\endbibitem

\bibitem[\protect\citeauthoryear{Gerlach~M}{2019}]{Gerlach2019}
\begin{barticle}
\bauthor{\bsnm{Gerlach~M}, \binits{A.L.} \bsuffix{Shi~H}}:
\batitle{A universal information theoretic approach to the identification of stopwords}.
\bjtitle{Nature Machine Intelligence}
\bvolume{1},
\bfpage{606}--\blpage{612}
(\byear{2019})
\doiurl{10.1038/s42256-019-0112-6}
\end{barticle}
\endbibitem

\bibitem[\protect\citeauthoryear{Kwon et~al.}{2023}]{kwon2023efficient}
\begin{bchapter}
\bauthor{\bsnm{Kwon}, \binits{W.}},
\bauthor{\bsnm{Li}, \binits{Z.}},
\bauthor{\bsnm{Zhuang}, \binits{S.}},
\bauthor{\bsnm{Sheng}, \binits{Y.}},
\bauthor{\bsnm{Zheng}, \binits{L.}},
\bauthor{\bsnm{Yu}, \binits{C.H.}},
\bauthor{\bsnm{Gonzalez}, \binits{J.E.}},
\bauthor{\bsnm{Zhang}, \binits{H.}},
\bauthor{\bsnm{Stoica}, \binits{I.}}:
\bctitle{Efficient memory management for large language model serving with pagedattention}.
In: \bbtitle{Proceedings of the ACM SIGOPS 29th Symposium on Operating Systems Principles}
(\byear{2023})
\end{bchapter}
\endbibitem

\bibitem[\protect\citeauthoryear{Cohen et~al.}{2009}]{cohen-2009-pearson}
\begin{botherref}
\oauthor{\bsnm{Cohen}, \binits{I.}},
\oauthor{\bsnm{Huang}, \binits{Y.}},
\oauthor{\bsnm{Chen}, \binits{J.}},
\oauthor{\bsnm{Benesty}, \binits{J.}},
\oauthor{\bsnm{Benesty}, \binits{J.}},
\oauthor{\bsnm{Chen}, \binits{J.}},
\oauthor{\bsnm{Huang}, \binits{Y.}},
\oauthor{\bsnm{Cohen}, \binits{I.}}:
Pearson correlation coefficient.
Noise reduction in speech processing,
1--4
(2009)
\end{botherref}
\endbibitem

\end{thebibliography}

\end{document}